\documentclass[preprint,12pt]{elsarticle}

\usepackage{amssymb}
\usepackage{amsmath}
\usepackage{graphicx}
\usepackage{caption}
\usepackage{subcaption}
\usepackage{booktabs}
\usepackage{multirow}
\usepackage{url}
\usepackage{hyperref}
\usepackage{lineno}
\usepackage{float}
\usepackage{tcolorbox}
\usepackage{enumitem}
\usepackage{placeins}
\usepackage{comment}

\usepackage{algorithm}
\usepackage{algpseudocode}

\usepackage{booktabs}    
\usepackage{longtable}   
\usepackage{array}       
\usepackage{multirow}    
\usepackage{siunitx}     

\usepackage{tikz}
\usepackage{pgfplots}
\pgfplotsset{compat=1.18}

\usepackage{enumitem}

\journal{}

\begin{document}

\begin{frontmatter}

\title{ES-C51: Expected Sarsa Based C51 Distributional Reinforcement Learning Algorithm}

\author[inst1]{Rijul Tandon\corref{cor1}}
\ead{letscomerijul@gmail.com}

\author[inst2]{Peter Vamplew}
\ead{p.vamplew@federation.edu.au}

\author[inst2]{Cameron Foale}
\ead{c.foale@federation.edu.au}

\address[inst1]{UIET, Panjab University , Chandigarh , India}
\address[inst2]{Federation University, Victoria, Australia}

\cortext[cor1]{Corresponding author}

\begin{abstract}
In most value-based reinforcement learning (RL) algorithms, the agent estimates only the expected reward for each action and selects the action with the highest reward. In contrast, Distributional Reinforcement Learning (DRL) estimates the entire probability distribution of possible rewards, providing richer information about uncertainty and variability. C51 is a popular DRL algorithm for discrete action spaces. It uses a Q-learning approach, where the distribution is learned using a greedy Bellman update. However, this can cause problems if multiple actions at a state have similar expected reward but with different distributions, as the algorithm may not learn a  stable distribution. This study presents a modified version of C51 (ES-C51) that replaces the greedy Q-learning update with an Expected Sarsa update, which uses a softmax calculation to combine information from all possible actions at a state rather than relying on a single best action. This reduces instability when actions have similar expected rewards and allows the agent to learn higher-performing policies. This approach is evaluated on classic control environments from Gym, and  Atari-10 games. For a fair comparison, we modify the standard C51's exploration strategy from $\epsilon$-greedy to softmax, which we refer to as QL-C51 (Q- Learning based C51). The results demonstrate that ES-C51 outperforms QL-C51 across many environments. 
\end{abstract}


\end{frontmatter}


\section{Introduction}

RL has demonstrated significant potential in training autonomous agents to operate effectively in stochastic environments. In standard RL, the goal is to learn an optimal policy $\pi(a|s)$ that maximizes the expected return. The action value function $Q(s,a)$ is given by equation \ref{eq:Action_Value_function}
\begin{equation}
\label{eq:Action_Value_function}
Q(s,a) = \mathbb{E} \left[ G \,\middle|\, s_0 = s, a_0 = a \right],
\end{equation}
where $G = \sum_{t=0}^{\infty} \gamma^t R_t$ is the return (cumulative discounted rewards starting from state $s$ and action $a$, following policy $\pi$),
$R_t$ and $\gamma \in (0,1]$ denote the reward and discounting factor at time step $t$. However, this approach only provides the expected reward $\mathbb{E}[G]$ for each state action pair $(s,a)$, without revealing the variability or uncertainty in outcomes. DRL addresses this issue by modeling the return distribution in equation \ref{eq:return_distribution} rather than just its expectation:
\begin{equation}
\label{eq:return_distribution}
Z^\pi(s,a) \sim P(G|s,a,\pi),
\end{equation}
 where $P(G|s,a,\pi)$ is the probability distribution over returns. This provides a complete view of the possible outcomes and their probabilities, enabling better decision making. One of the most widely used algorithms in DRL is C51, which approximates the distribution $Z^\pi(s,a)$ using a categorical distribution. C51 updates its distributions using a Q-learning based greedy policy shown by equation \ref{eq:distribution_update}  where the next distribution is chosen from the action with the highest expected reward given by equation \ref{eq:greedy_action}

\begin{equation}
\label{eq:distribution_update}
Z(s,a) = R(s,a) + \gamma \times Z(s',a').
\end{equation}

\begin{equation}
\label{eq:greedy_action}
a' = \arg\max_{a \in \mathcal{A}} \mathbb{E}[Z(s',a)].
\end{equation}

While effective in many cases, this greedy selection can fail when multiple actions have identical expected rewards but different variances. For example, consider a state $s$ where two possible actions $a_1$ and $a_2$ satisfy equation \ref{eq:equal_expected_reward}
\begin{equation}    
\label{eq:equal_expected_reward}
\mathbb{E}[Z(s,a_1)] = \mathbb{E}[Z(s,a_2)],
\end{equation}
where $\mathbb{E}[Z(s,a)]$ is given by equation \ref{eq:Expected_State_Action_Reward} where $p_\theta(s,a,z_i)$ represents the probability of atom $z_i$ for state action pair $(s,a)$ and $N$ denotes the number of atoms
\begin{equation} 
\label{eq:Expected_State_Action_Reward}
Q(s,a) = \mathbb{E}[Z(s,a)] = \sum_{i=0}^{N-1} z_i \cdot p_\theta(s,a,z_i) 
\end{equation}
yet their distributions differ in variance shown by equation \ref{eq:Uneven_variance}
\begin{equation}
\label{eq:Uneven_variance}
\mathrm{Var}(Z(s,a_1)) \ll \mathrm{Var}(Z(s,a_2)) 
\end{equation}
where variance is calculated via equation \ref{eq:variance_formula}
\begin{equation}  
\label{eq:variance_formula}
\text{Var}(Z(s,a)) = \sum_{i=0}^{N-1} p_\theta(s,a,z_i) (z_i - \mathbb{E}[Z(s,a)])^2
\end{equation}

\begin{figure}[H]
    \centering
    \includegraphics[width=0.6\linewidth]{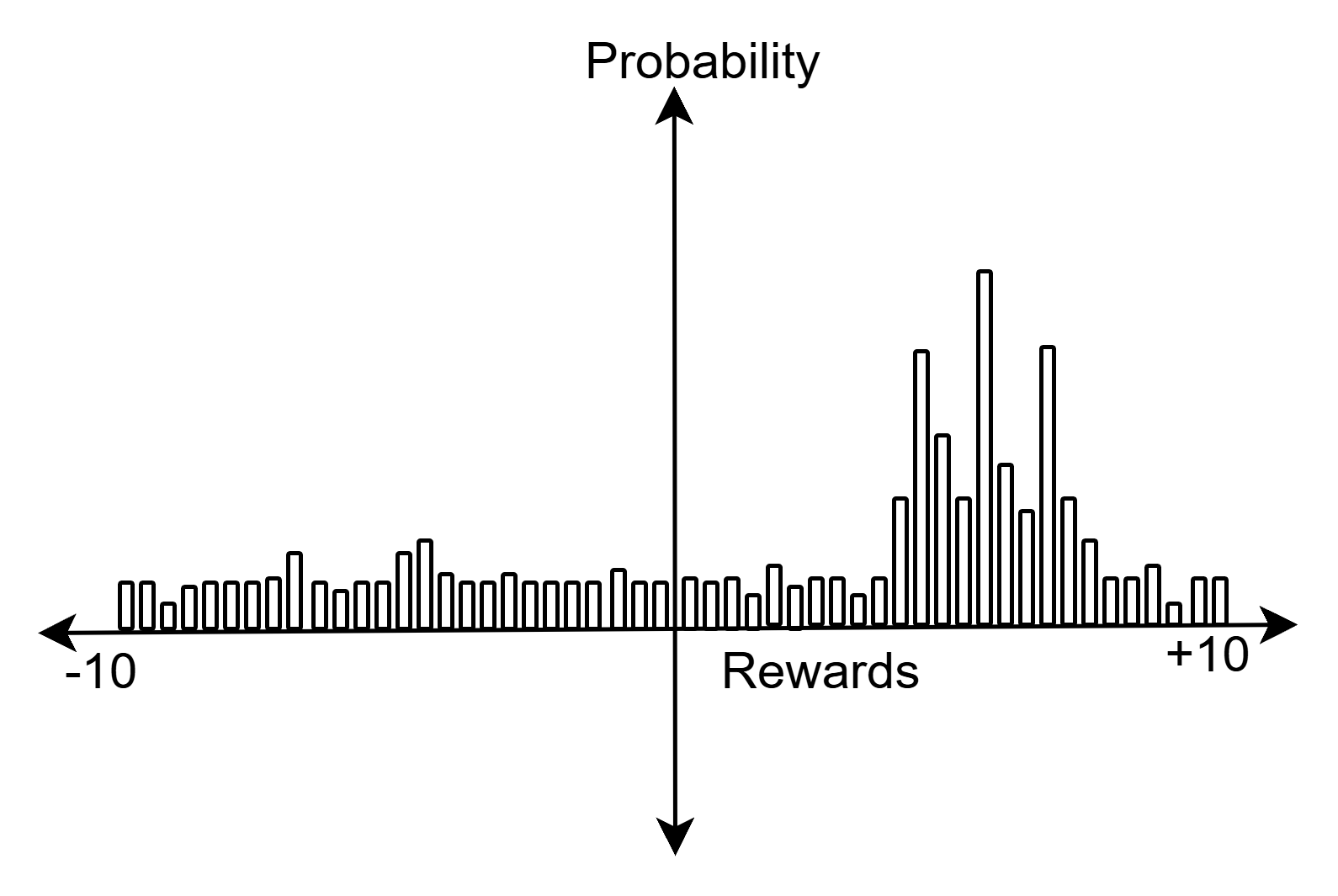}
    \caption{Example of a stable C51 distribution $Z(s,a_1)$ with low variance.}
    \label{fig:stable_graph}
\end{figure}

In such cases, an agent choosing $a_2$ might experience unstable performance due to high variance in returns. Figure~\ref{fig:stable_graph} illustrates  $Z(s,a_1)$  a stable  C51 reward distribution for a given state action pair $(s,a_1)$.

\begin{figure}[H]
    \centering
    \includegraphics[width=0.6\linewidth]{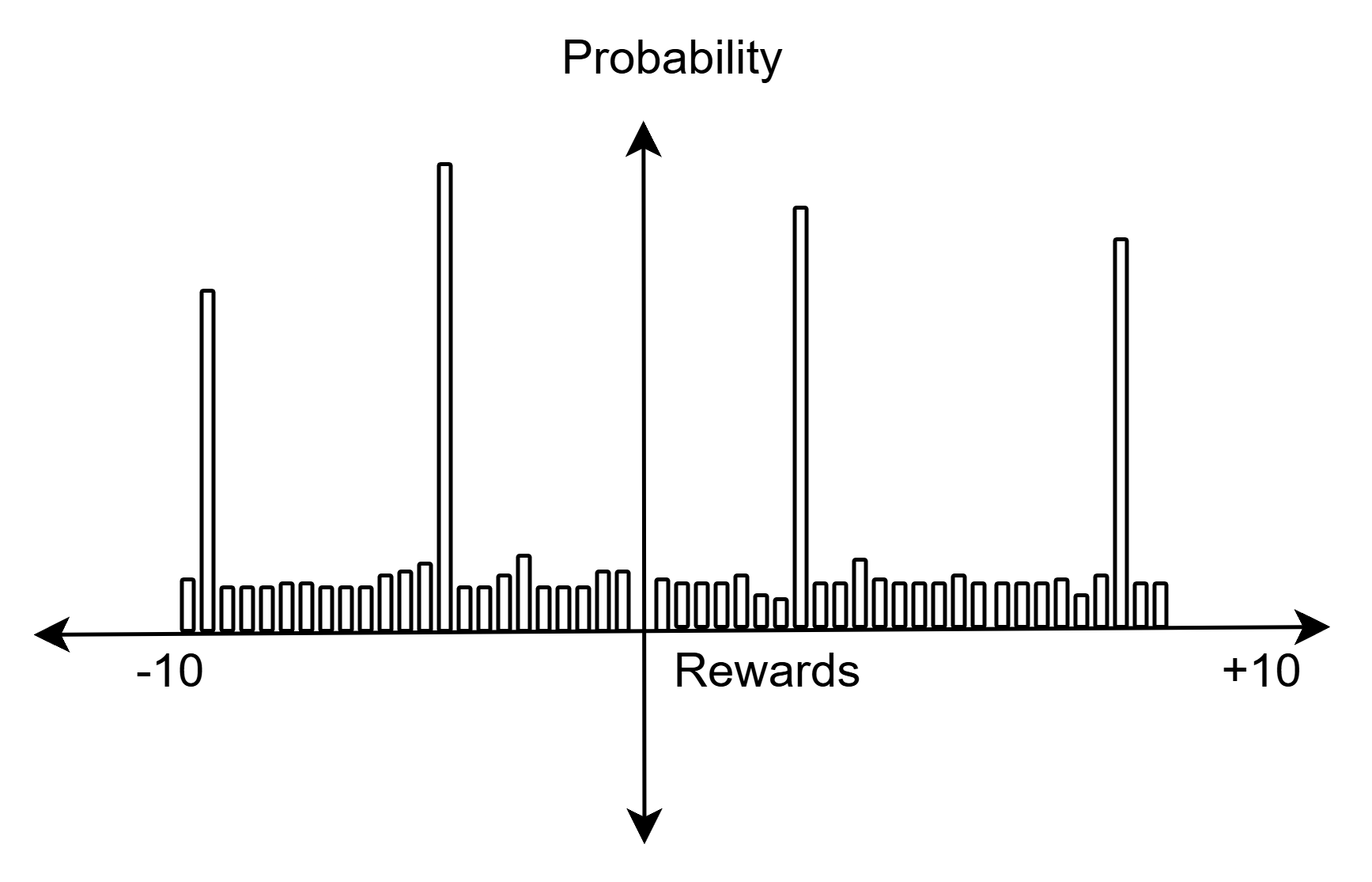}
    \caption{Example of an unstable C51 distribution $Z(s,a_2)$ with high variance.}
    \label{fig:unstable_graph}
\end{figure}

 Figure~\ref{fig:unstable_graph} illustrates $Z(s,a_2)$ an unstable, high variance distribution the selection of which can lead to poor long term performance even after the learning phase. 

Moreover an important source of instability arises in the value estimate for the action leading to $s_t$ from the previous state $(s_{t-1},a)$. Suppose the agent initially selects $a_1$ as the greedy action in $s_t$. Then $Z(s_{t-1},a)$ will converge towards a discounted version of $Z(s_t,a_1)$. However, the greedy action at $s_t$ may later switch, either due to $a_1$ 
sampling from the lower tail of its distribution or due to exploratory selection of $a_2$ which yields a high return. 
Although this change has little impact on $Q(s_t,a)$, since 
$Z(s_t,a_1)$ and $Z(s_t,a_2)$ share the same expectation, it can cause substantial updates to $Z(s_{t-1},a)$. 
As a result, repeated switches in greedy choice may delay the stabilization of $Z(s_{t-1},a)$. This intuition is similar to the value function interference problem identified in multi-objective reinforcement learning (MORL), where multiple values 
for each $(s,a)$ pair must be combined to estimate $Q(s,a)$ 
and greedy action switching introduces instability \cite{vamplew2024value}. It has been observed that due to `policy churn' greedy action selections are subject to rapid change in deep RL, with around 10\% of all states undergoing a change in greedy action after a single gradient update of DQN \cite{schaul2022phenomenon}. To address the possible instability produced in C51 by policy churn, this study proposes the ES-C51 algorithm which incorporates concepts from the Expected Sarsa algorithm \cite{van2009theoretical} into C51. Instead of relying solely on the greedy choice, ES-C51 updates the current distribution using a softmax over all possible actions. The main contributions of this paper are as follows:  

\begin{itemize}
    \item \textbf{Algorithmic modification:} Study extends the standard QL-C51 algorithm by introducing ES-C51 (Expected Sarsa based C51) variant.  
    
    \item \textbf{Policy and distributional updates:} Both C51 implementations were modified to employ a softmax based action selection strategy. In QL-C51, the Bellman backup remains greedy, whereas in ES-C51 the update leverages a softmax policy with a decaying temperature parameter ($\tau$). As $\tau$ gradually approaches zero, ES-C51 naturally converges to a greedy policy.  
    
    \item \textbf{Comprehensive evaluation:} Large scale experiments using 10 independent seeds across Atari-10 environments \cite{aitchison2023atari} (in both stochastic and deterministic variants) and two classic control Gym environments. The results were analyzed using the Wilcoxon signed rank test, and comparative performance plots were generated.  
    
    \item \textbf{Performance interpretation:} This study presents a comprehensive quantitative evaluation of performance across all environments, including a detailed time efficiency comparison between both approaches. These quantify the performance gains achievable with ES-C51.
\end{itemize}

\section{Literature Survey}
\subsection{Distributional reinforcement learning (DRL)}
DRL marks a pivotal advance in the field of RL, where instead of estimating the expected return for state–action pairs, agents learn the full probability distribution of returns. This approach addresses core limitations of scalar RL, which collapses all reward information into a single expected value thereby losing crucial details about risk sensitivity and variance in returns. \cite{rowland2019statistics}  presents a unifying framework for designing and analysing DRL algorithms in terms of recursively estimating statistics of the return distribution. Bellemare et al. \cite{bellemare2017distributional} introduced the C51 algorithm, the first practical DRL approach, mapping returns to a categorical distribution over a fixed set of atoms using a novel projection operator. Rowland et al.~\cite{rowland2018analysis} provided a theoretical analysis of the projected distributional Bellman operator, relating it to the Cramér distance metric and giving convergence guarantees for sample-based categorical algorithms. Kastner et al.~\cite{kastner2025categorical} later extended this analysis to the more practical KL-divergence loss, closing the gap between theory and practice. 

C51 inspired further advances, such as quantile-based approaches like QR-DQN and Implicit Quantile Networks (IQN) \cite{dabney2018distributional, dabney2018implicit} supplanted atom based methods by directly learning quantile functions or continuous quantile parameterizations. These works highlight the value of flexible distributional parameterization, but also reveal practical challenges, such as instability and computational cost. 

Recent refinements, such as Dual Expectile Quantile Regression (DEQR)~\cite{jullien2025distributional}, show that hybrid statistical measures can yield gains in stability and efficiency. Enwerem et al. \cite{enwerem2025safety} proposes a risk-regularized quantile-based algorithm integrating Conditional Value-at-Risk (CVaR) to enforce safety without complex architectures and provide theoretical guarantees on the contraction properties of the risk-sensitive distributional Bellman operator in Wasserstein space, ensuring convergence to a unique cost distribution. Shaik et al.~\cite{shaik2025generalized} introduced Distributional Generalized Advantage Estimation, leveraging the Wasserstein like metric for robust policy gradient methods. DRL concepts now underpin generalizations to multi-agent and multi-objective RL. Hu et al.~\cite{hu2022distributional} developed a distributional reward estimation framework for more stable multi-agent learning. Hayes et al. \cite{hayes2023monte} combines distributional methods with Monte-Carlo Tree Search for multi-objective problems. Eldeeb et al.~\cite{eldeeb2024conservative} demonstrates the use of DRL with conservative Q-learning for offline multi-agent tasks under uncertainty. 

\subsection{On-policy and Off-policy Temporal Difference Updates}

Deep Q-Networks (DQN) marked a breakthrough in reinforcement learning by achieving human-level performance in Atari games using raw pixels \cite{mnih2015human}. As an off-policy extension of Q-learning, DQN learns by evaluating the best theoretical next action and stabilizes training with a target network. Sarsa, by contrast, is an on-policy method that updates based on the actual action taken, making it sensitive to exploratory behavior and often yielding safer policies \cite{rummery1994line}, but also creating variance in updates. Expected Sarsa sits between the two: it averages updates over all possible next actions under the current policy, reducing variance and improving stability \cite{Sutton2018}. Unlike Expected Sarsa, DQN’s maximization can cause value overestimation, which Double DQN addresses \cite{van2016deep}. Expected Sarsa naturally mitigates this issue by averaging. \cite{van2009theoretical} proves that Expected Sarsa converges under the same conditions as Sarsa and formulate specific hypotheses about when Expected Sarsa will outperform Sarsa and Q-learning. Experiments in multiple domains confirm these hypotheses and demonstrate that Expected Sarsa has significant advantages over these more commonly used methods. Despite these theoretical and empirical benefits Expected Sarsa remains underexplored in DRL. This gap motivates our investigation into Expected Sarsa-based DRL algorithms, where the combination of expectation-based updates and distributional value representations may offer complementary advantages in stability and sample efficiency.

\section{Proposed Approach: ES-C51 vs. QL-C51}
This study proposes ES-C51, a modification of the standard QL-C51 algorithm.
 We chose C51 over other forms of DRL as the basis for our experiments as it is one of the most widely used and was compatible with the proposed Expected Sarsa modifications, moreover an easily modifiable implementation was available via CleanRL \citep{huang2022cleanrl}. The key innovation in ES-C51 is the use of a softmax weighted expectation over next actions in the Bellman backup, rather than the greedy action used in QL-C51. This change aims to improve exploration and stabilize learning, especially in the early stages of training. In DRL the goal is to model the full distribution of returns, $Z(s,a)$ for each state action pair $(s,a)$, rather than just the expected return. The return distribution is represented as a vector which gives probabilities of getting each return value and the probabilities are calculated by a neural network parameterized by $\theta$ denoted by $p_\theta(s,a,z_i)$. Thus, the distribution is represented by equation  \ref{eq:distribution}. For action selection, both QL-C51 and ES-C51 employ a softmax policy over Q-values, parameterized by a temperature $\tau$ as shown in equation \ref{eq:softmax_policy}
\begin{equation} 
\label{eq:softmax_policy}
\pi_\tau(a \mid s) = \frac{\exp(Q(s,a)/\tau)}{\sum_{b\in\mathcal{A}} \exp(Q(s,b)/\tau)}
\end{equation} 
where $\mathcal{A}$ is the set of all possible actions. The temperature $\tau$ controls the exploration exploitation trade off: high values of $\tau$ result in a nearly uniform policy (more exploration), while low values make the policy more greedy. In implementation both ES-C51 and QL-C51 use a softmax policy where an action is sampled from $\pi_\tau$. The main difference between QL-C51 and ES-C51 lies in the construction of the target distribution for the temporal difference (TD) update. In QL-C51, the TD target is constructed using the distribution corresponding to the greedy action in the next state. Specifically, after computing $Q(s,a)$ for all $a \in \mathcal{A}$, the greedy action $a'$ is selected as shown in equation \ref{eq:greedy_action}. The distributional Bellman operator under policy $\pi$ is given by equation \ref{eq:distribution_bellman_operator} 
 \begin{equation} 
\label{eq:distribution_bellman_operator}
\mathcal{T}^\pi Z(s,a) \overset{D}{=} R(s,a) + \gamma \times Z(s',a'), \quad a' \sim \pi(\cdot\mid s')
\end{equation} where $R(s,a)$ is the immediate reward received after taking action $a$ in state $s$, $s'$ is the next state, and $a'$ is sampled from the policy $\pi$ at $s'$. The notation $\overset{D}{=}$ denotes equality in distribution. The C51 algorithm approximates the return distribution $Z(s,a)$ using a categorical distribution supported on a fixed set of $N$ atoms $\{z_i \}_{i=0}^{N-1}$, which are uniformly spaced in the interval $[v_{\min}, v_{\max}]$. For every atom $z_i$ in the categorical support, equation \ref{eq:Each_atom_update} computes its shifted value $v_i$, reflecting how the distribution’s return atoms are updated after each transition.
\begin{equation} 
\label{eq:Each_atom_update}
v_i = r(s,a) + \gamma\, z_i, \quad i \in [0, N-1]
\end{equation}

where $r(s,a)$ is the observed reward. Since the shifted atoms $v_i$ will, in general, not exactly align with the fixed atom support $\{z_i\}$ the projection operator $\phi$ ( refer to equation 7 in \cite{bellemare2017distributional}) is applied. It redistributes the probability mass from each $v_i$ onto the nearest atoms in the fixed support using interpolation resulting in the target distribution \ref{eq:distribution}, The result is a probability vector over the original atom support forming $Z_{target}$ for the (s,a) pair.
\begin{equation}
\label{eq:distribution}
\begin{gathered}
Z_{\text{target}} =\phi\Big(Z(s,a)\Big) \\
Z_{target}= Z(s,a)= \big\{ p_\theta(s, a, z_0),\, p_\theta(s, a, z_1),\, \dots,\, p_\theta(s, a, z_{N-1}) \big\}
\end{gathered}
\end{equation}

In contrast, ES-C51 constructs the TD target as the expectation over all possible next actions, weighted by their probabilities under the softmax policy shown by equation \ref{eq:softmax_policy}. The distributional Bellman operator under policy $\pi_\tau$  for expected sarsa update is given by Equation \ref{eq:distribution_bellman_operator_expected_sarsa} 
 \begin{equation} 
\label{eq:distribution_bellman_operator_expected_sarsa}
\mathcal{T}^\pi Z(s,a) \overset{D}{=} R(s,a) +  \gamma \sum_{a' \in \mathcal{A}} \pi_\tau(a'|s') \cdot Z(s',a')
\end{equation} 

This distribution is shifted using equation \ref{eq:Each_atom_update}
and projected onto the fixed support using equation \ref{eq:distribution}. In both algorithms, the neural network is trained by minimizing the cross entropy loss between the predicted distribution and the projected target distribution: 
\begin{equation} 
\mathcal{L}(\theta) = - \sum_{i=0}^{N-1} Z_{\text{target}}[i]\log p_\theta(s,a,z_i).
\end{equation} 
 To summarize, both QL-C51 and ES-C51 share the same softmax exploration mechanism, the same projection operator $\phi$, and the same cross entropy training objective. Their only difference lies in the TD backup: QL-C51 projects the distribution of the greedy action, while ES-C51 projects the expectation of the softmax weighted mixture of next action distributions as shown on Line 14 of Algorithm \ref{alg:esC51}. This modification in ES-C51 is designed to reduce variance and improve stability, especially during early learning when Q-value estimates are noisy.


\begin{table}[h]
\centering
\caption{Notations}
\label{tab:notations}
\scriptsize
\begin{tabular}{ll|ll}
\hline
\textbf{Symbol} & \textbf{Meaning} & \textbf{Symbol} & \textbf{Meaning} \\ \hline
$s, s'$ & Current state, next state & $a, a'$ & Current action, next action \\
$r(s,a)$ & Immediate reward& $\gamma$ & Discount factor \\
$\mathcal{A}$ & Action space & $\mathcal{D}$ & Replay buffer storing transitions $(s,a,r,s')$ \\
$N$ & Number of atoms & $\{z_i\}_{i=0}^{N-1}$ & Fixed atom support in $[v_{\min}, v_{\max}]$ \\
$v_{\min}, v_{\max}$ & Min and max atom support values & $v_i$ & Shifted atom value: $r(s,a) + \gamma z_i$ \\
$p_{\theta}(s,a,z_i)$ & Probability of atom $z_i$ for $(s,a)$ & $f_\theta(s,a)$ & Network logits for $(s,a)$ \\
$p_{\theta}$ & Neural network  & $p_{\theta}$ & Target network copy of $p_\theta$ \\
$Z_\theta(s,a)$ & Return distribution for $(s,a)$ & $Z_{\theta'}(s,a)$ & Return distribution from target network \\
$\bar{Z}(s')$ & Expected mixture distribution & $Q(s,a)$ & Expected return: $\sum_i z_i p_\theta(s,a,z_i)$ \\
$\pi_\tau(a|s)$ & Softmax policy with temperature $\tau$ & $\tau$ & Temperature parameter, decaying to $0$ \\
$\mathcal{T}^\pi$ & Distributional Bellman operator & $\mathcal{L}(\theta)$ & Cross entropy loss function \\
$Z_{\text{target}}$ & Target distribution after projection & $\theta, \theta'$ & Network and target parameters \\
$t, T$ & Current timestep, total timesteps & $\phi$ & Projection Operator \\
\hline
\end{tabular}
\end{table}

\begin{algorithm}
\caption{ES-C51}
\label{alg:esC51}
\begin{algorithmic}[1]
\State Initialize replay buffer $\mathcal{D}$, parameters $\theta$ of network $p_\theta$, and target network $p_{\theta'} \gets p_\theta$
\State Initialize fixed support of atoms $\{z_i\}_{i=0}^{N-1}$ uniformly in $[v_{\min}, v_{\max}]$
\For{each timestep $t = 1, \dots, T$}
    \State Observe state $s_t$
    \State Compute logits $f_\theta(s_t, a)$ and probability masses $p_\theta(s_t, a, z_i)$ \Statex \hspace*{\algorithmicindent}for all $a \in \mathcal{A}$ and $z_i$
    \State Compute action values $Q(s_t,a) = \sum_{i=0}^{N-1} z_i \, p_\theta(s_t, a, z_i)$
    \State Select action $a_t \sim \pi_{\tau(\cdot|s_t)}$ where
    \Statex \hspace*{\algorithmicindent}\hspace*{\algorithmicindent} $\pi_{\tau(a \mid s_t)} = \frac{\exp\big(Q(s_t,a)/\tau\big)}{\sum_{b \in \mathcal{A}} \exp\big(Q(s_t,b)/\tau\big)}$
    \State Execute $a_t$, observe reward $r_t$ and next state $s_{t+1}$
    \State Store transition $(s_t, a_t, r_t, s_{t+1})$ in $\mathcal{D}$
    \If{$t >$ learning\_starts and $t \bmod$ train\_frequency $=0$}
        \State Sample mini batch $(s, a, r, s')$ from $\mathcal{D}$
        \State Compute $Q(s', a') = \sum_{i=0}^{N-1} z_i \, p_{\theta'}(s', a', z_i)$ for all $a'$
        \State Compute $\pi_{\tau}(a'|s')$ as above
        \State Compute expected distribution for next state-action pairs:
        \Statex \hspace*{\algorithmicindent}\hspace*{\algorithmicindent} $\bar{Z}(s') = \sum_{a' \in \mathcal{A}} \pi_\tau(a'|s') \cdot Z(s', a')$
        \State For each atom $z_i$ from $\bar{Z}(s')$, compute shifted values $v_i = r(s,a) + \gamma z_i$ onto fixed support $\{z_i\}$ using  $\phi$ to obtain $Z_{target}$
        \State Update $\theta$ by minimizing cross entropy loss:
        \State \hspace{0.5cm} $\mathcal{L}(\theta) = - \sum_{i=0}^{N-1} Z_{\text{target}}[i]\log p_\theta(s,a,z_i)$
    \EndIf
    \If{$t \bmod$ target\_update $= 0$}
        \State Update target network: $\theta' \gets \theta$
    \EndIf
\EndFor
\end{algorithmic}
\end{algorithm}

\section{Experimental Setup and Implementation Details}

All experiments were conducted on the OzSTAR supercomputing facility at Swinburne University of Technology, using NVIDIA P100 GPUs and single CPU cores. We evaluated our algorithms on Gym classic control and Atari environments, testing each game in both deterministic NoFrameskip-v4 and stochastic v0 settings. Two distributional reinforcement learning algorithms were compared: the baseline QL-C51 and our modified ES-C51. Both algorithms were set to use softmax based action selection, with QL-C51 employing a greedy Bellman backup, uses an Expected Sarsa backup with the probability of actions determined by the same softmax operation and temperature $\tau$ used for action selection as detailed in Algorithm \ref{alg:esC51}. Each environment algorithm pair was trained across 10 random seeds, yielding a total of 440 independent runs. Training was managed with CleanRL, and statistical comparisons were conducted using the Wilcoxon signed rank test. Results were additionally grouped into stochastic and deterministic categories to highlight the influence of environment dynamics. Performance plots were generated by averaging across seeds, applying smoothing for clarity, and plotting mean returns with confidence intervals for both algorithms on the same axes. All standard hyperparameters specified in CleanRL were used for 
implementing both the Atari and Gym versions of ES-C51. In addition we introduced a parameter $\tau$ to enable a softmax policy. 
This parameter is initialized with a value of $1$ and decays according to equation \ref{eq:tau_decay}
\begin{equation}
\label{eq:tau_decay}
\tau = \max\!\left(1.0 \times \left(1 - \frac{\text{Current\_Timestep}}{\text{Total\_Timesteps*0.75}}\right),\, 0.01\right)
\end{equation}

\section{Results}
\subsection {Summary of results} 
\begin{figure}[H]
    \centering
    \includegraphics[width=\linewidth]{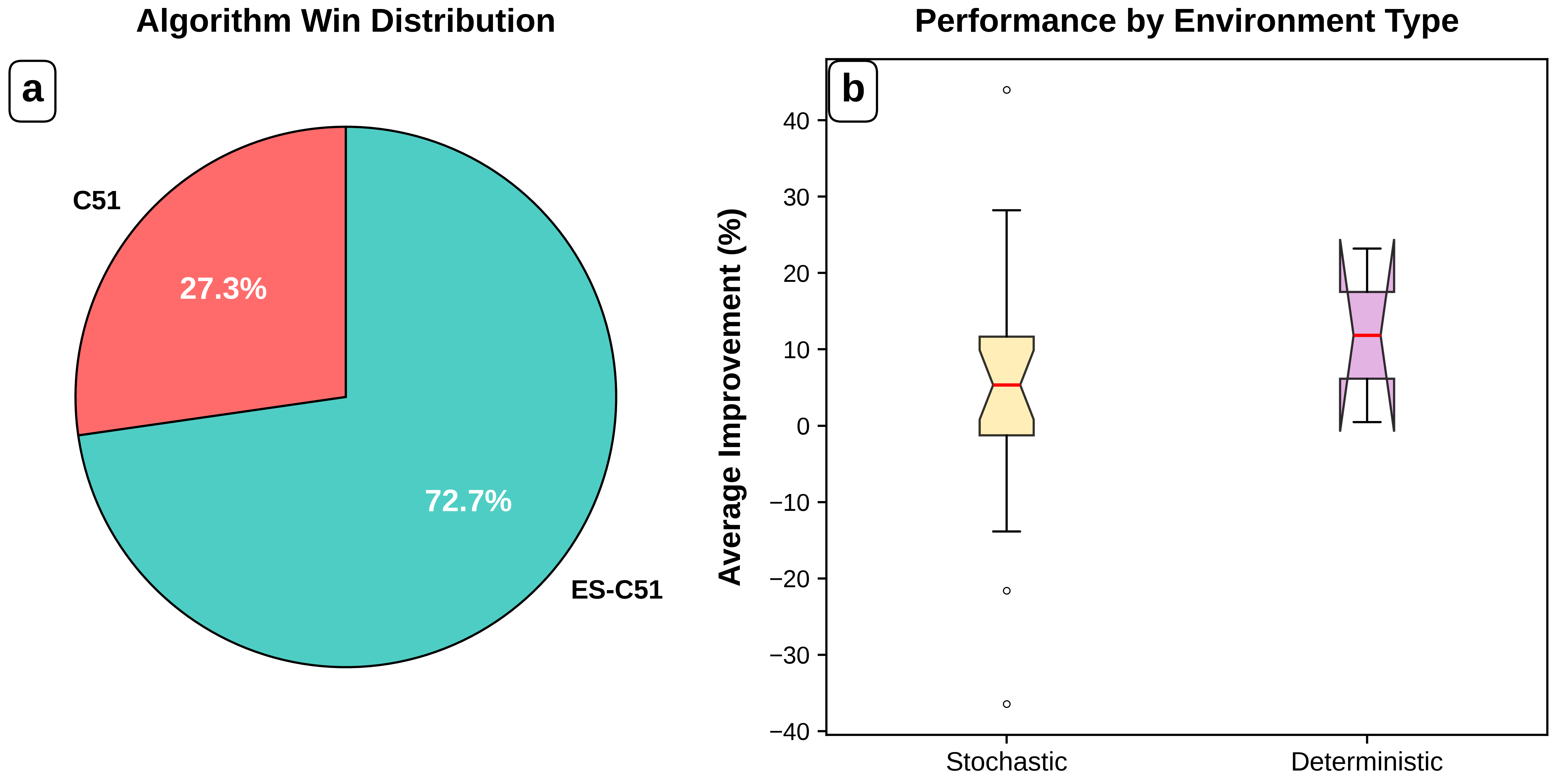}
    \caption{Performance summary}
    \label{fig:performance_summary}
\end{figure}

\begin{figure}
    \centering
    \includegraphics[width=\linewidth]{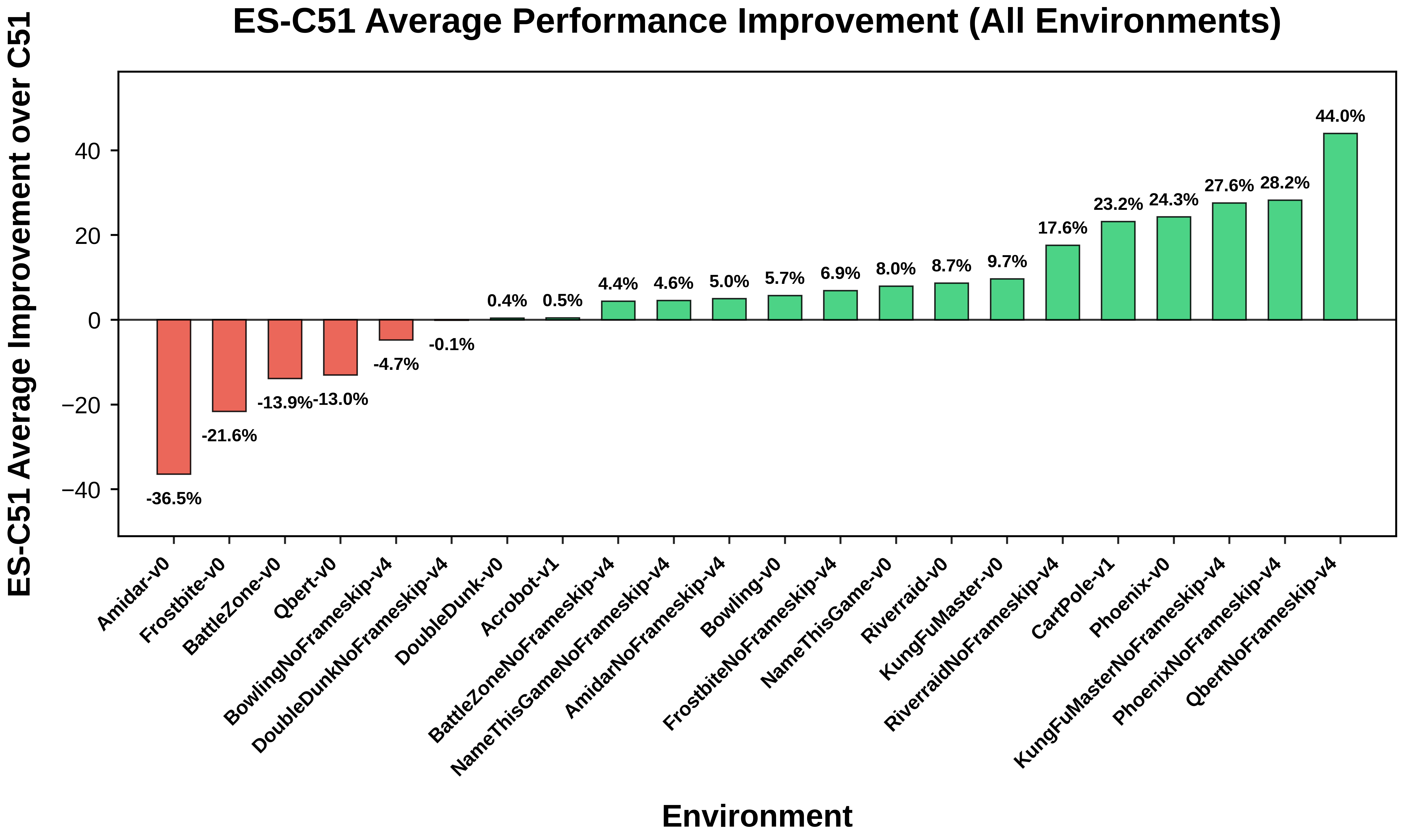}
    \caption{Average Performance Improvement by ES-C51}
    \label{fig:Average performance improvement}
\end{figure}

Table \ref{tab:runtime-summary} provides the runtime comparison of both ES-C51 and QL-C51. Despite the extra calculations required by ES-C51, for some environments it takes less execution time when trained for the same number of timesteps. ES-C51 may have more streamlined gradient updates or reduced overhead in certain parts of the learning process that offset the additional ES computations. Table \ref{tab:environment-results} , and Figures \ref{fig:performance_summary} and \ref{fig:Average performance improvement} summarise the overall performance of ES-C51 and QL-C51 across all environments. Figure \ref{fig:Average performance improvement}  shows a visual analysis of improvement in every environment while Table \ref{tab:environment-results} shows the exact value of mean rewards and standard deviation along with the \% improvements. It considers only the final 10\% of episodes from each training run, as this segment reflects the agent’s mean performance after reaching a stable, fully trained state. This approach ensures that the reported results represent the agent’s proficiency at the culmination of the learning process, minimizing the influence of early-stage variability. Pie chart  \ref{fig:performance_summary} (a) shows that ES-C51 achieves a higher mean reward than QL-C51 on 72.7\% of the environments. Box plot \ref{fig:performance_summary}  (b) illustrates the range of improvements made by ES-C51 on stochastic and deterministic environments. Notably, deterministic environments demonstrate more consistent and superior performance with improvements ranging from -4.74\% to 43.98\%, with a median around 6-7\% and most values concentrated between 4\% and 28\%. In contrast, stochastic environments show more varied and volatile results, ranging from -36.46\% to 24.27\%, with a median near 7\% but exhibiting greater spread and several negative outliers, indicating that ES-C51's performance is less stable in stochastic settings compared to deterministic ones.

\begin{table}[h]
\centering
\caption{Runtime Comparison Analysis grouped by environment type}
\label{tab:runtime-summary}
\scriptsize
\begin{tabular}{l l S[table-format=2.1] 
                S[table-format=5.1] 
                S[table-format=5.1]}
\toprule
\textbf{Env.} & \textbf{Winner} & {\textbf{Imp. (\%)}} 
& {\textbf{QL-C51 (s)}} & {\textbf{ES-C51 (s)}} \\
\midrule

\multicolumn{5}{l}{\textbf{Gym Environments}} \\
\midrule
Acrobot-v1      & ES-C51 &  0.4  & 1017.3 & 1013.1 \\
CartPole-v1     & ES-C51 & 22.5  & 1111.1 &  907.2 \\

\midrule
\multicolumn{5}{l}{\textbf{Atari Environments (Stochastic)}} \\
\midrule
Amidar-v0        & ES-C51 &  4.2  & 64230.9 & 61669.6 \\
BattleZone-v0    & QL-C51                  & -6.9  & 67790.8 & 72839.9 \\
Bowling-v0       & QL-C51                  & -7.0  & 56172.7 & 60400.6 \\
DoubleDunk-v0    & ES-C51 & 13.4  & 68264.6 & 60206.5 \\
Frostbite-v0     & ES-C51 &  5.9  & 70243.7 & 66319.0 \\
KungFuMaster-v0  & QL-C51                  & -1.6  & 63538.8 & 64559.8 \\
NameThisGame-v0  & QL-C51                  & -0.7  & 63246.8 & 63685.4 \\
Phoenix-v0       & QL-C51                  & -0.2  & 54065.8 & 54155.7 \\
Riverraid-v0     & ES-C51 &  5.1  & 70503.0 & 67095.6 \\

\midrule
\multicolumn{5}{l}{\textbf{Atari Environments (Deterministic, NoFrameskip)}} \\
\midrule
AmidarNoFrameskip-v4        & QL-C51                  & -7.4  & 48386.8 & 52270.9 \\
BattleZoneNoFrameskip-v4    & QL-C51                  & -3.0  & 47953.2 & 49416.5 \\
BowlingNoFrameskip-v4       & QL-C51                  & -3.8  & 46701.2 & 48525.6 \\
DoubleDunkNoFrameskip-v4    & ES-C51 &  5.7  & 54619.1 & 51685.8 \\
FrostbiteNoFrameskip-v4     & QL-C51                  & -0.1  & 50065.2 & 50114.7 \\
KungFuMasterNoFrameskip-v4  & ES-C51 &  9.2  & 50889.1 & 46618.2 \\
NameThisGameNoFrameskip-v4  & QL-C51                  & -1.8  & 52615.5 & 53604.0 \\
PhoenixNoFrameskip-v4       & ES-C51 &  5.7  & 43458.5 & 41110.9 \\
QbertNoFrameskip-v4         & ES-C51 &  1.6  & 51539.3 & 50713.8 \\
RiverraidNoFrameskip-v4     & QL-C51                  & -1.5  & 48504.1 & 49217.8 \\

\bottomrule
\end{tabular}
\end{table}

\begin{table}[h]
\centering
\caption{Detailed Environment Results grouped by environment type}
\label{tab:environment-results}
\scriptsize
\begin{tabular}{l l S[table-format=3.2] S[table-format=1.4] l l l}
\toprule
\textbf{Env.} & \textbf{Winner} & {\textbf{Imp. (\%)}} & {\textbf{P-val.}} & \textbf{Sig.} & \textbf{QL-C51 M $\pm$ S} & \textbf{ES-C51 M $\pm$ S} \\
\midrule
\multicolumn{7}{l}{\textbf{Gym Environments}} \\
\midrule
Acrobot-v1 & ES-C51 & 0.47 & 0.5566 & False & -253.63 $\pm$ 199.72 & -252.45 $\pm$ 201.04 \\
CartPole-v1 & ES-C51 & 23.20 & 0.0020 & True & 377.07 $\pm$ 17.68 & 464.55 $\pm$ 12.00 \\
\midrule
\multicolumn{7}{l}{\textbf{Atari Stochastic Environments}} \\
\midrule
DoubleDunk-v0 & ES-C51 & 0.40 & 0.0371 & True & -21.20 $\pm$ 0.06 & -21.12 $\pm$ 0.09 \\
KungFuMaster-v0 & ES-C51 & 9.68 & 0.1309 & False & 1263.45 $\pm$ 221.39 & 1385.74 $\pm$ 206.32 \\
Bowling-v0 & ES-C51 & 5.71 & 0.0020 & True & 29.06 $\pm$ 0.60 & 30.71 $\pm$ 0.49 \\
Qbert-v0 & QL-C51 & -13.02 & 0.3750 & False & 270.57 $\pm$ 49.37 & 235.34 $\pm$ 104.49 \\
Riverraid-v0 & ES-C51 & 8.65 & 0.0195 & True & 1800.43 $\pm$ 89.73 & 1956.19 $\pm$ 128.39 \\
Frostbite-v0 & QL-C51 & -21.60 & 0.0645 & False & 109.50 $\pm$ 16.97 & 85.84 $\pm$ 24.08 \\
Phoenix-v0 & ES-C51 & 24.27 & 0.0039 & True & 800.54 $\pm$ 66.35 & 994.84 $\pm$ 166.91 \\
BattleZone-v0 & QL-C51 & -13.86 & 0.0137 & True & 3625.91 $\pm$ 325.37 & 3123.54 $\pm$ 204.38 \\
Amidar-v0 & QL-C51 & -36.46 & 0.0098 & True & 19.12 $\pm$ 6.39 & 12.15 $\pm$ 5.18 \\
NameThisGame-v0 & ES-C51 & 7.95 & 0.1055 & False & 2502.24 $\pm$ 401.33 & 2701.29 $\pm$ 338.07 \\
\midrule
\multicolumn{7}{l}{\textbf{Atari Deterministic (NoFrameskip) Environments}} \\
\midrule
RiverraidNoFrameskip-v4 & ES-C51 & 17.60 & 0.0020 & True & 1883.08 $\pm$ 123.96 & 2214.53 $\pm$ 95.50 \\
NameThisGameNoFrameskip-v4 & ES-C51 & 4.56 & 0.0840 & False & 2677.40 $\pm$ 132.45 & 2799.62 $\pm$ 175.13 \\
BowlingNoFrameskip-v4 & QL-C51 & -4.74 & 0.0137 & True & 27.22 $\pm$ 0.67 & 25.93 $\pm$ 0.87 \\
PhoenixNoFrameskip-v4 & ES-C51 & 28.22 & 0.0098 & True & 1946.82 $\pm$ 357.10 & 2496.25 $\pm$ 271.67 \\
KungFuMasterNoFrameskip-v4 & ES-C51 & 27.56 & 0.0039 & True & 3316.21 $\pm$ 662.74 & 4230.29 $\pm$ 1019.56 \\
FrostbiteNoFrameskip-v4 & ES-C51 & 6.90 & 0.3750 & False & 109.07 $\pm$ 15.28 & 116.59 $\pm$ 20.42 \\
BattleZoneNoFrameskip-v4 & ES-C51 & 4.40 & 0.0840 & False & 2434.12 $\pm$ 137.04 & 2541.15 $\pm$ 149.08 \\
AmidarNoFrameskip-v4 & ES-C51 & 4.98 & 1.0000 & False & 3.02 $\pm$ 1.07 & 3.17 $\pm$ 1.72 \\
DoubleDunkNoFrameskip-v4 & QL-C51 & -0.11 & 0.8457 & False & -18.04 $\pm$ 0.21 & -18.06 $\pm$ 0.17 \\
QbertNoFrameskip-v4 & ES-C51 & 43.98 & 0.9219 & False & 273.82 $\pm$ 117.77 & 394.24 $\pm$ 428.28 \\
\bottomrule
\end{tabular}
\end{table}

\FloatBarrier

\subsection{Analysis of Performance on Individual Environments}
\begin{figure}[H]
    \centering
    \begin{subfigure}[b]{0.48\linewidth}
        \includegraphics[width=\linewidth]{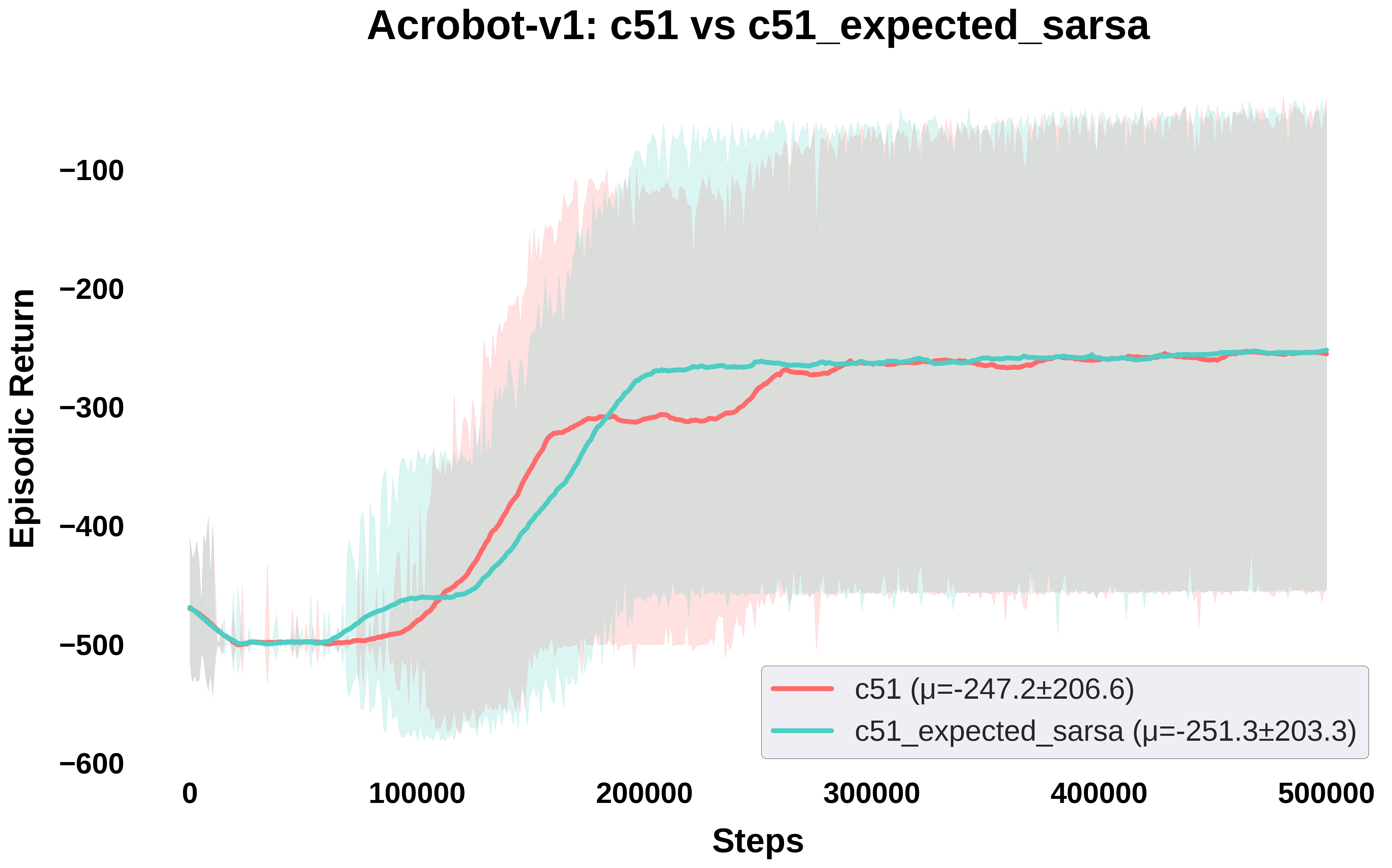}
        \caption{Acrobot-v1}
        \label{fig:acrobot_v1}
    \end{subfigure}
    \hfill
    \begin{subfigure}[b]{0.48\linewidth}
        \includegraphics[width=\linewidth]{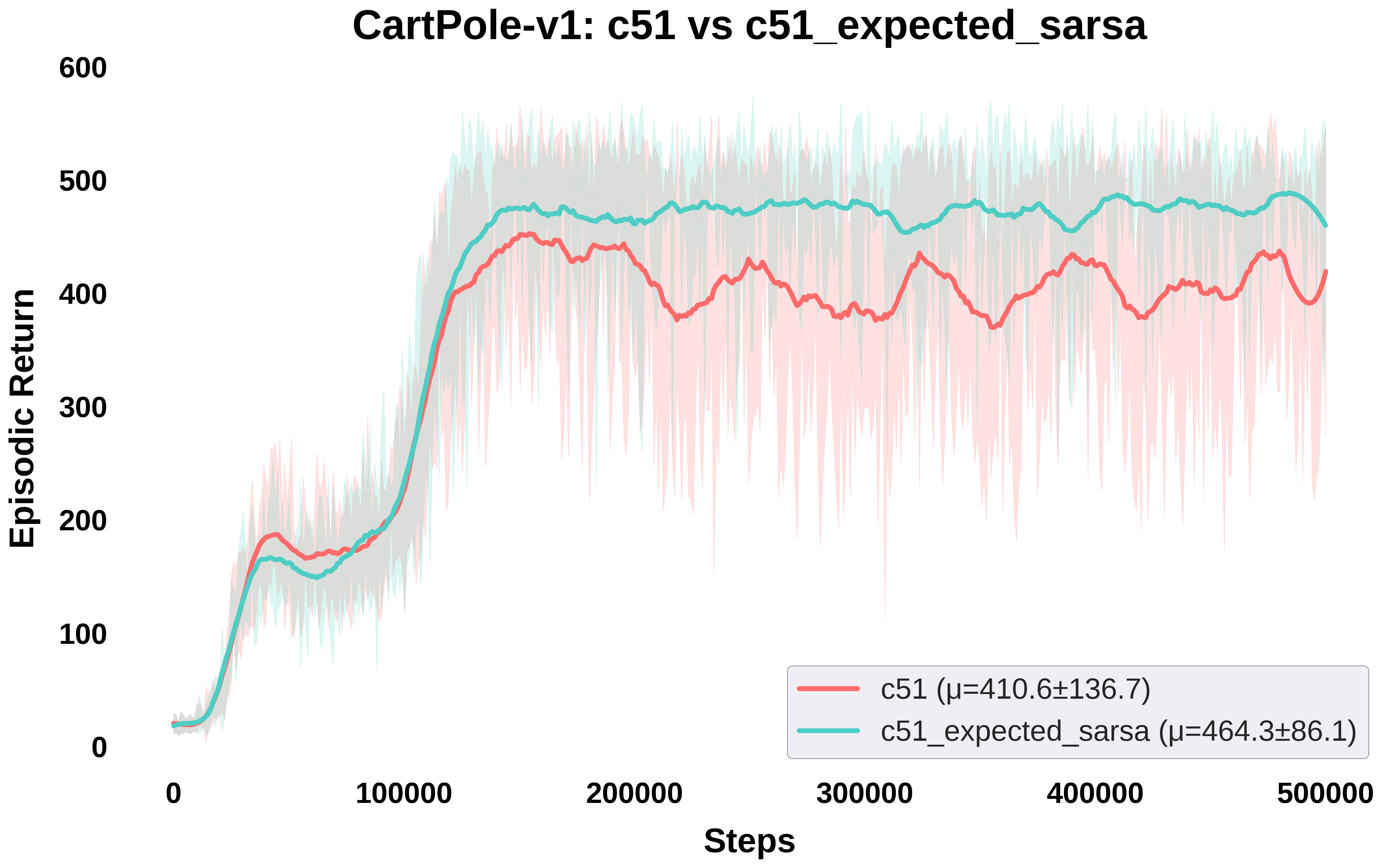}
        \caption{CartPole-v1}
        \label{fig::Cartpole_v1}
    \end{subfigure}
    \caption{Performance on Gym Classic Control Environments.}
    \label{fig:gym_comparison}
\end{figure}

Figure \ref{fig:gym_comparison} shows the performance comparison of both algorithms on Acrobot and Cartpole classic control gym environments. Figure \ref{fig:acrobot_v1}  shows that ES-C51 converged to global optimal reward at around 20000 steps while QL-C51 took 30000 steps to do the same . Moreover ES-C51 shows better stability with less fluctuations .
Figure \ref{fig::Cartpole_v1} shows that in Cartpole ES-C51 achieves a large jump in performance as compared to QL-C51, ES-C51 converges to a reward of around 500 while QL-C51 fluctuates around 400.

\begin{figure}[H]
    \centering
    \begin{subfigure}[b]{0.48\linewidth}
        \includegraphics[width=\linewidth]{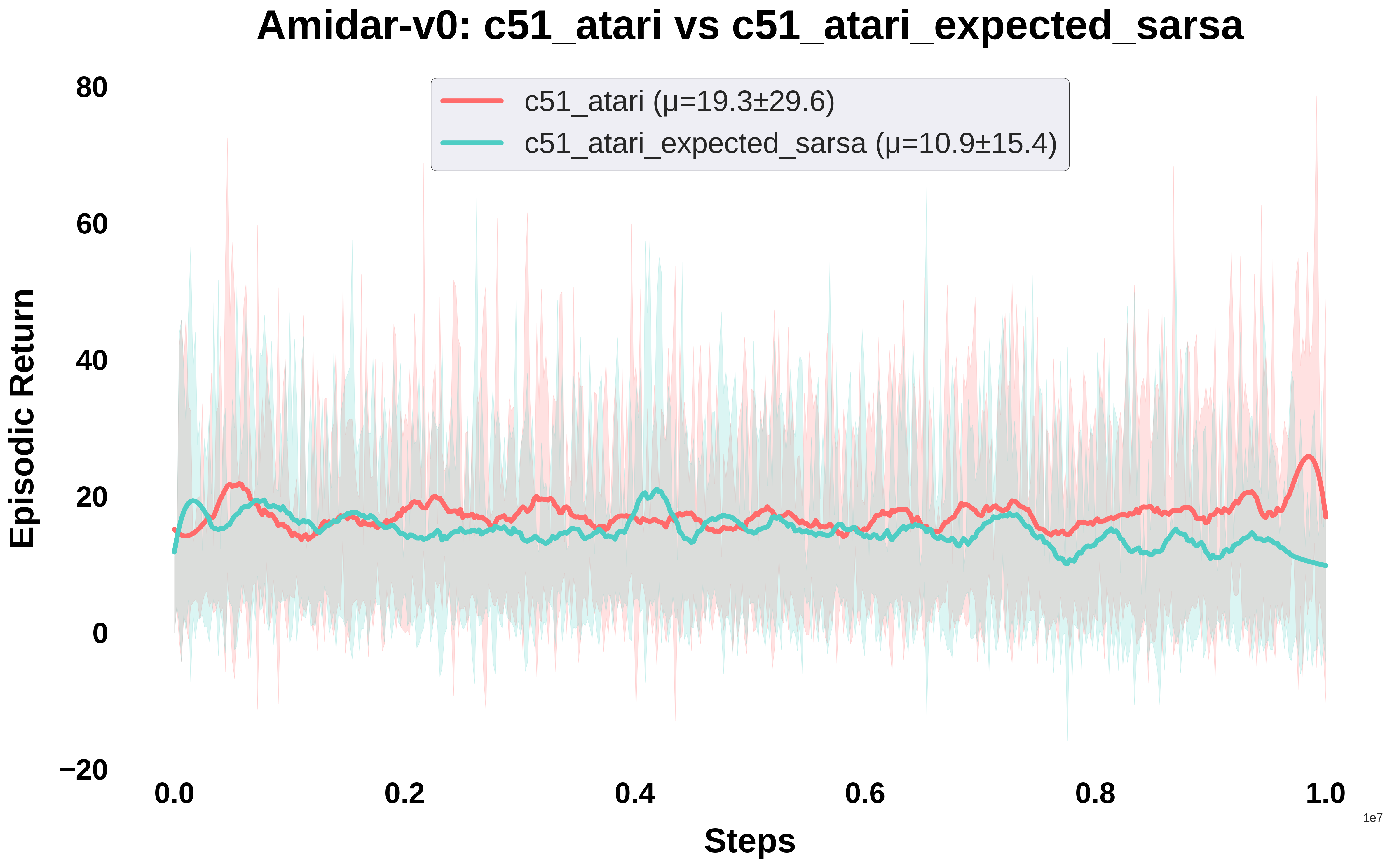}
        \caption{Amidar-v0}
        \label{fig:Amidar_v0}
    \end{subfigure}
    \hfill
    \begin{subfigure}[b]{0.48\linewidth}
        \includegraphics[width=\linewidth]{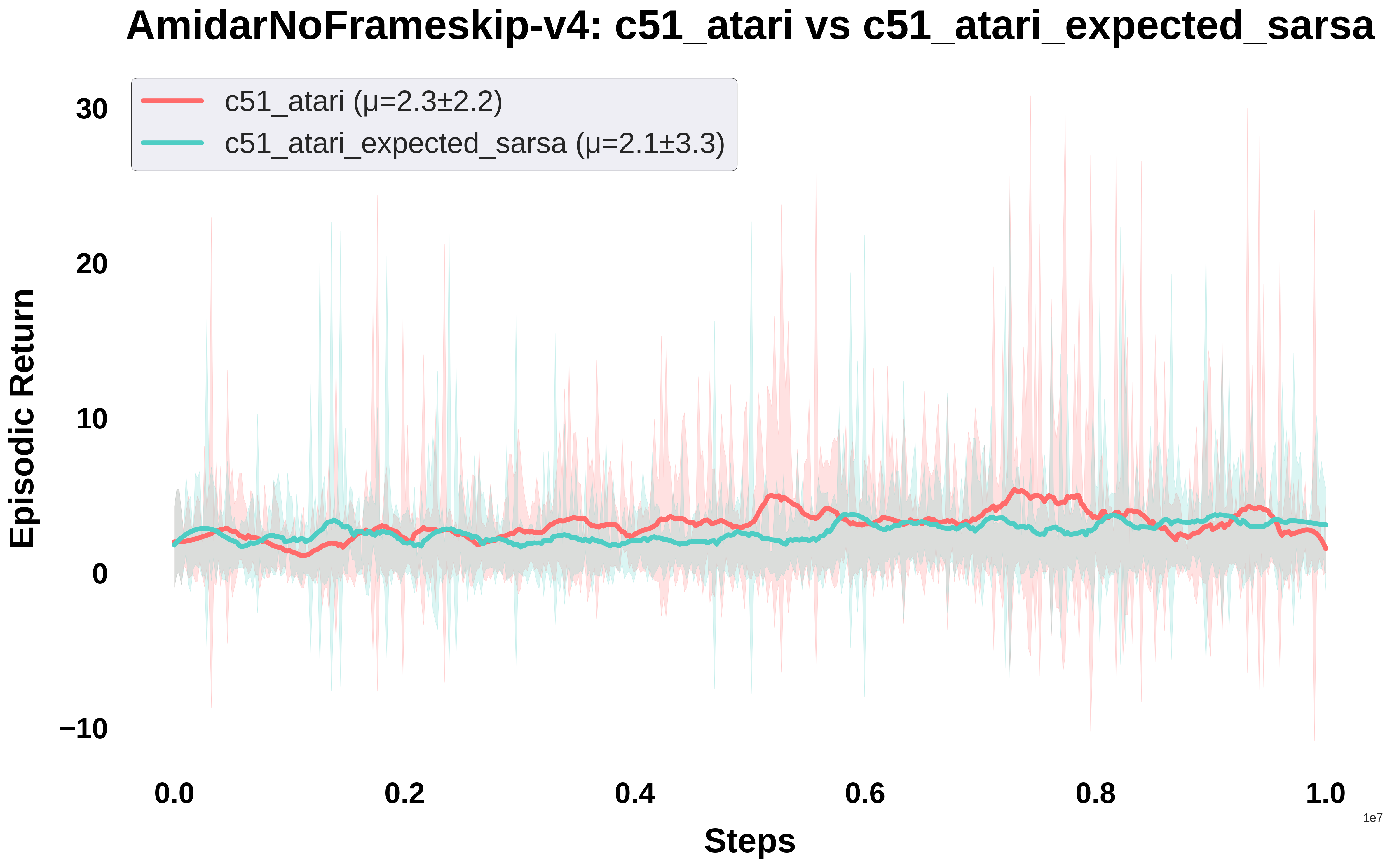}
        \caption{AmidarNoFrameskip-v4}
        \label{fig:Amidar_no_frameskip}
    \end{subfigure}
    \caption{Comparison results for Amidar environments.}
    \label{fig:Amidar_pair}
\end{figure}

\begin{figure}[H]
    \centering
    \begin{subfigure}[b]{0.48\linewidth}
        \includegraphics[width=\linewidth]{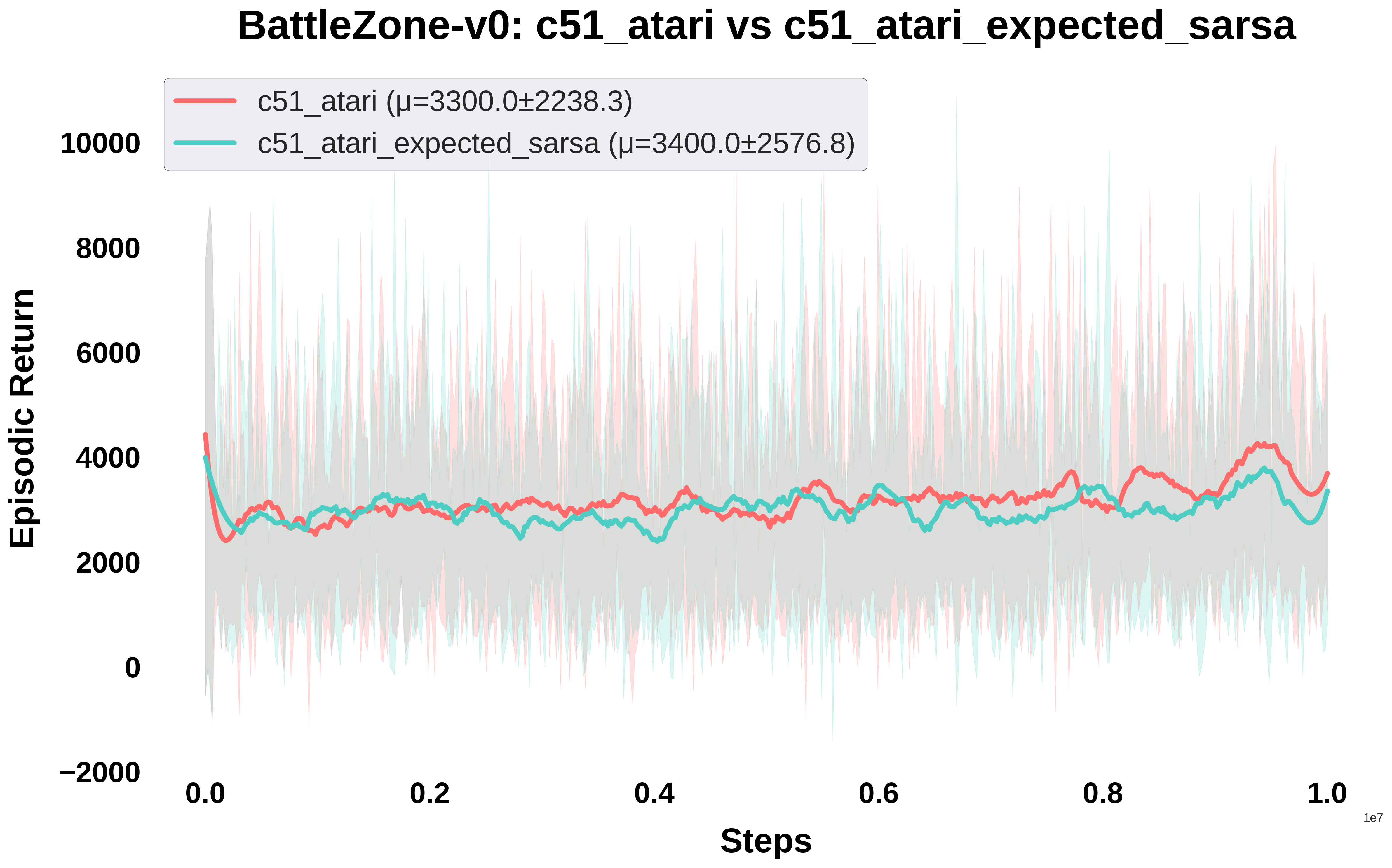}
        \caption{BattleZone-v0}
        \label{fig:battlezone_v0}
    \end{subfigure}
    \hfill
    \begin{subfigure}[b]{0.48\linewidth}
        \includegraphics[width=\linewidth]{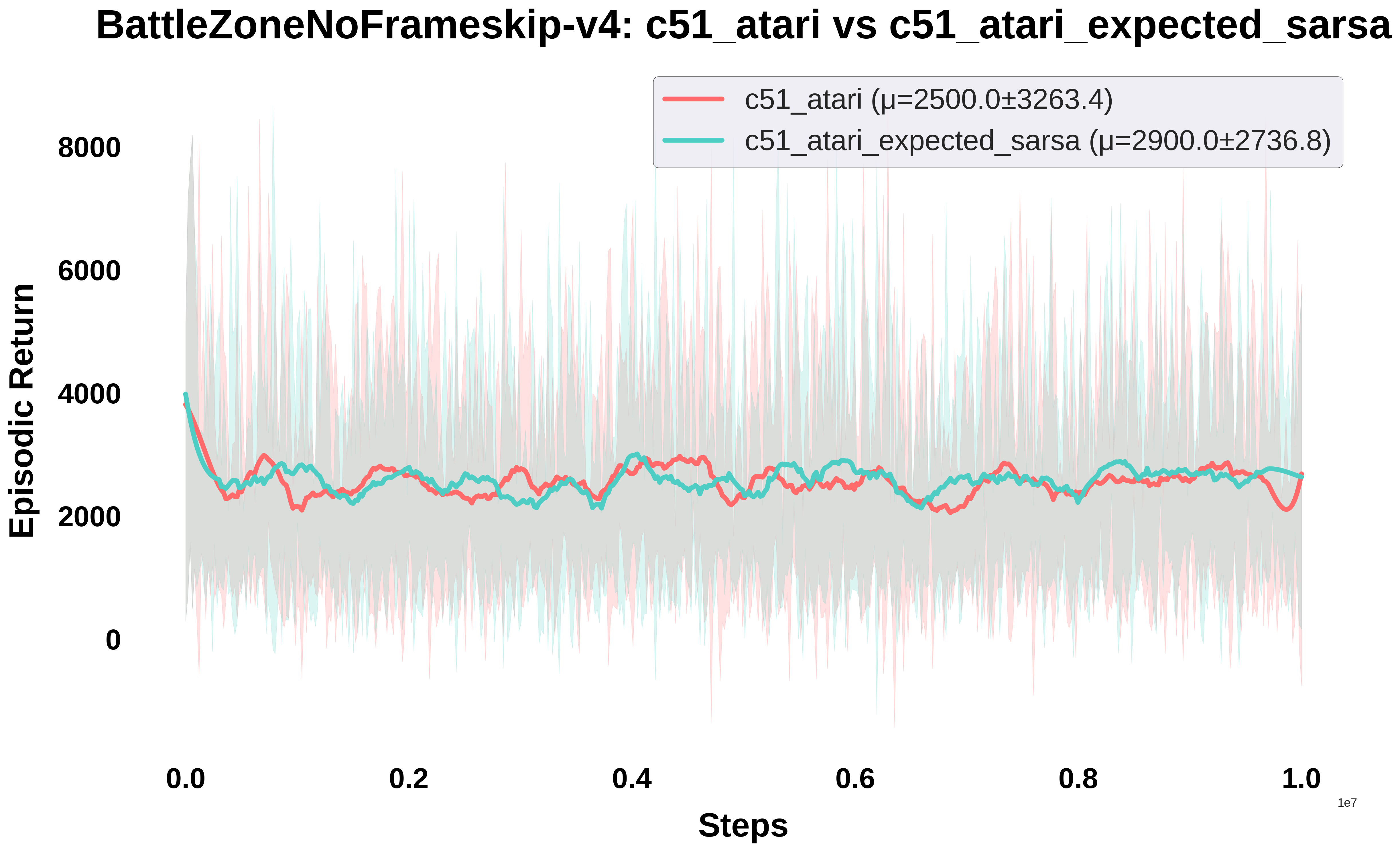}
        \caption{BattleZoneNoFrameskip-v4}
        \label{fig:battlezone_no_frameskip}
    \end{subfigure}
    \caption{Comparison results for BattleZone environments.}
    \label{fig:battlezone_pair}
\end{figure}

\begin{figure}[H]
    \centering
    \begin{subfigure}[b]{0.48\linewidth}
        \includegraphics[width=\linewidth]{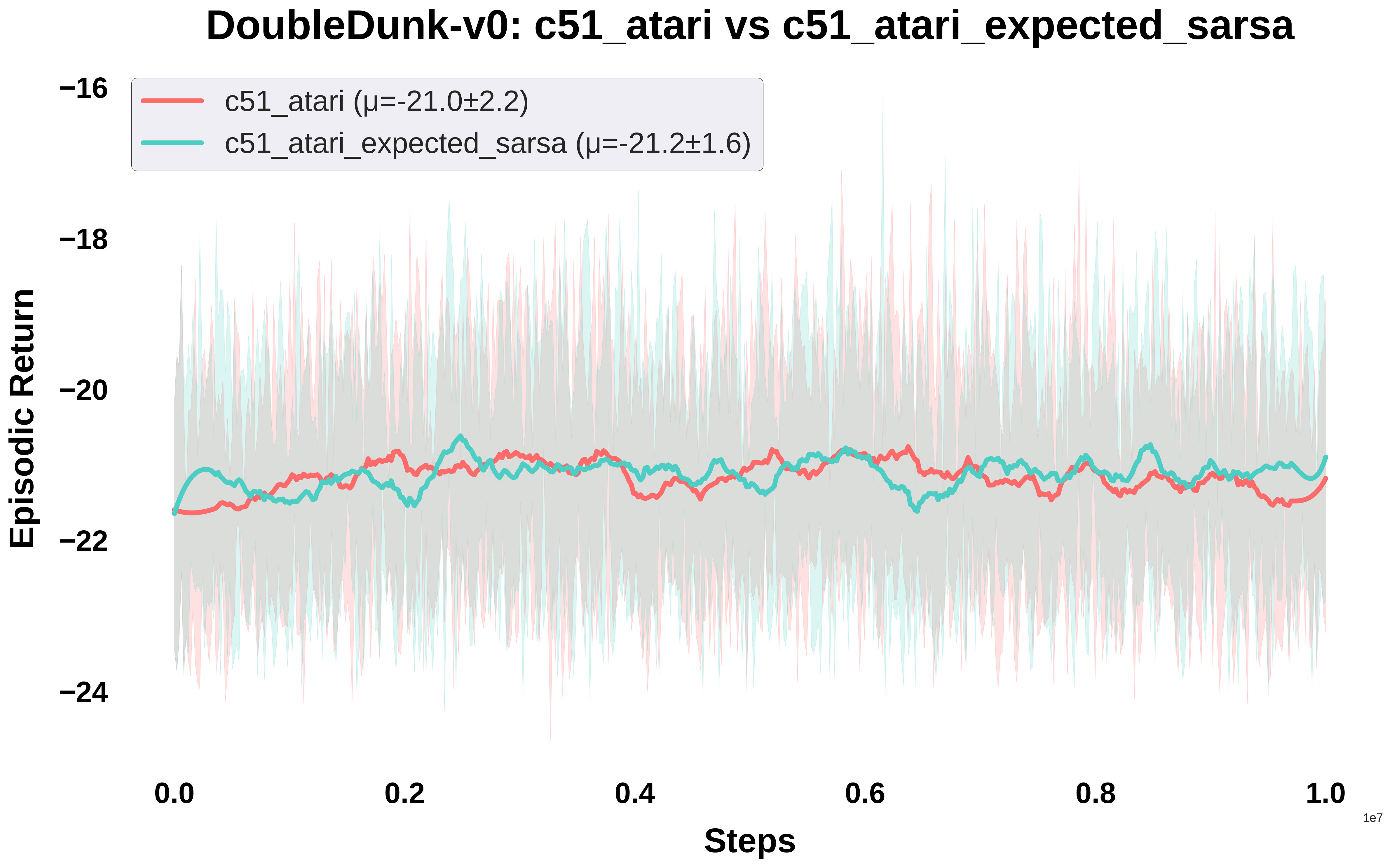}
        \caption{DoubleDunk-v0}
        \label{fig:doubledunk_v0}
    \end{subfigure}
    \hfill
    \begin{subfigure}[b]{0.48\linewidth}
        \includegraphics[width=\linewidth]{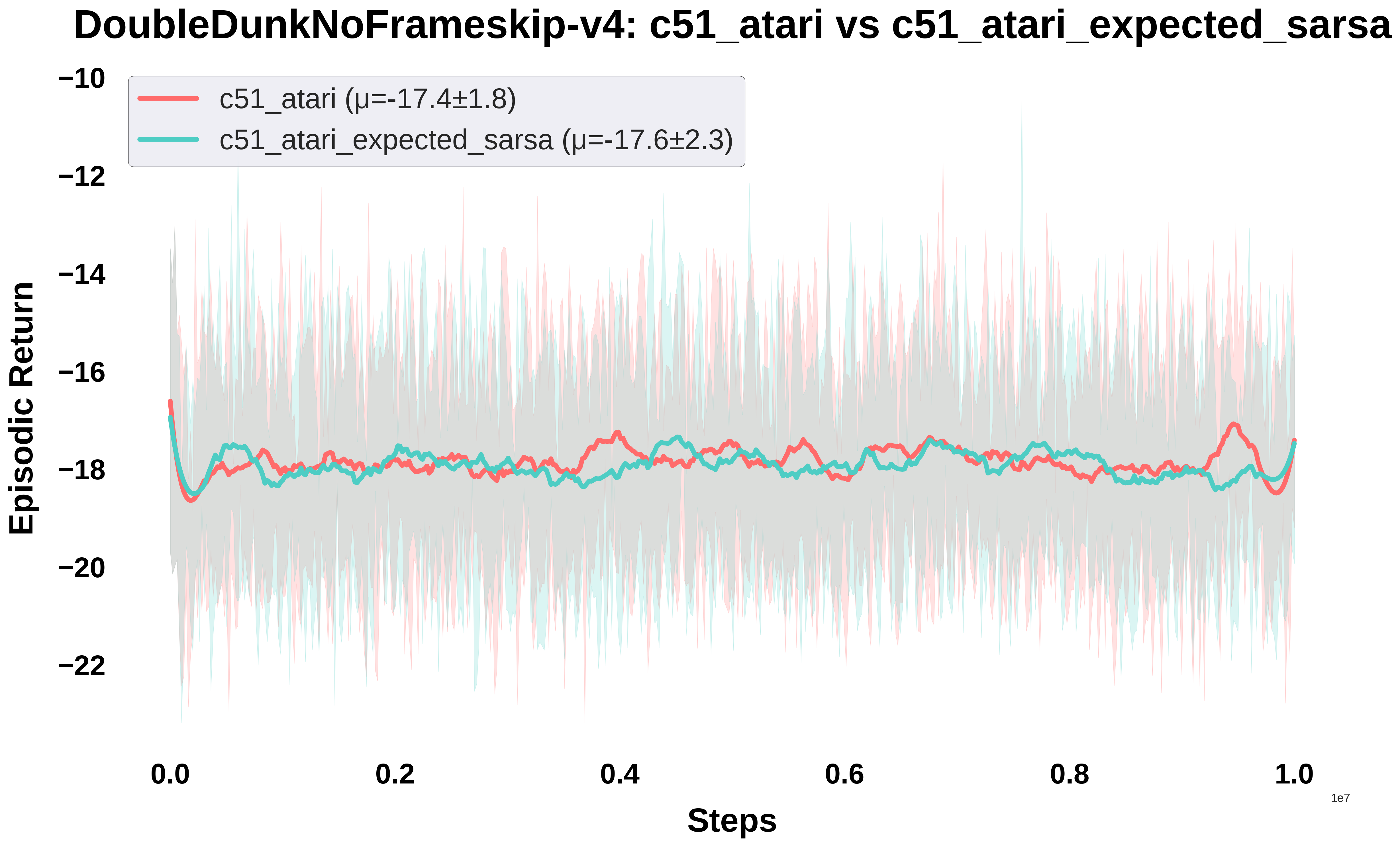}
        \caption{DoubleDunkNoFrameskip-v4}
        \label{fig:doubledunk_no_frameskip}
    \end{subfigure}
    \caption{Comparison results for DoubleDunk environments.}
    \label{fig:doubledunk_pair}
\end{figure}

\begin{figure}[H]
    \centering
    \begin{subfigure}[b]{0.48\linewidth}
        \includegraphics[width=\linewidth]{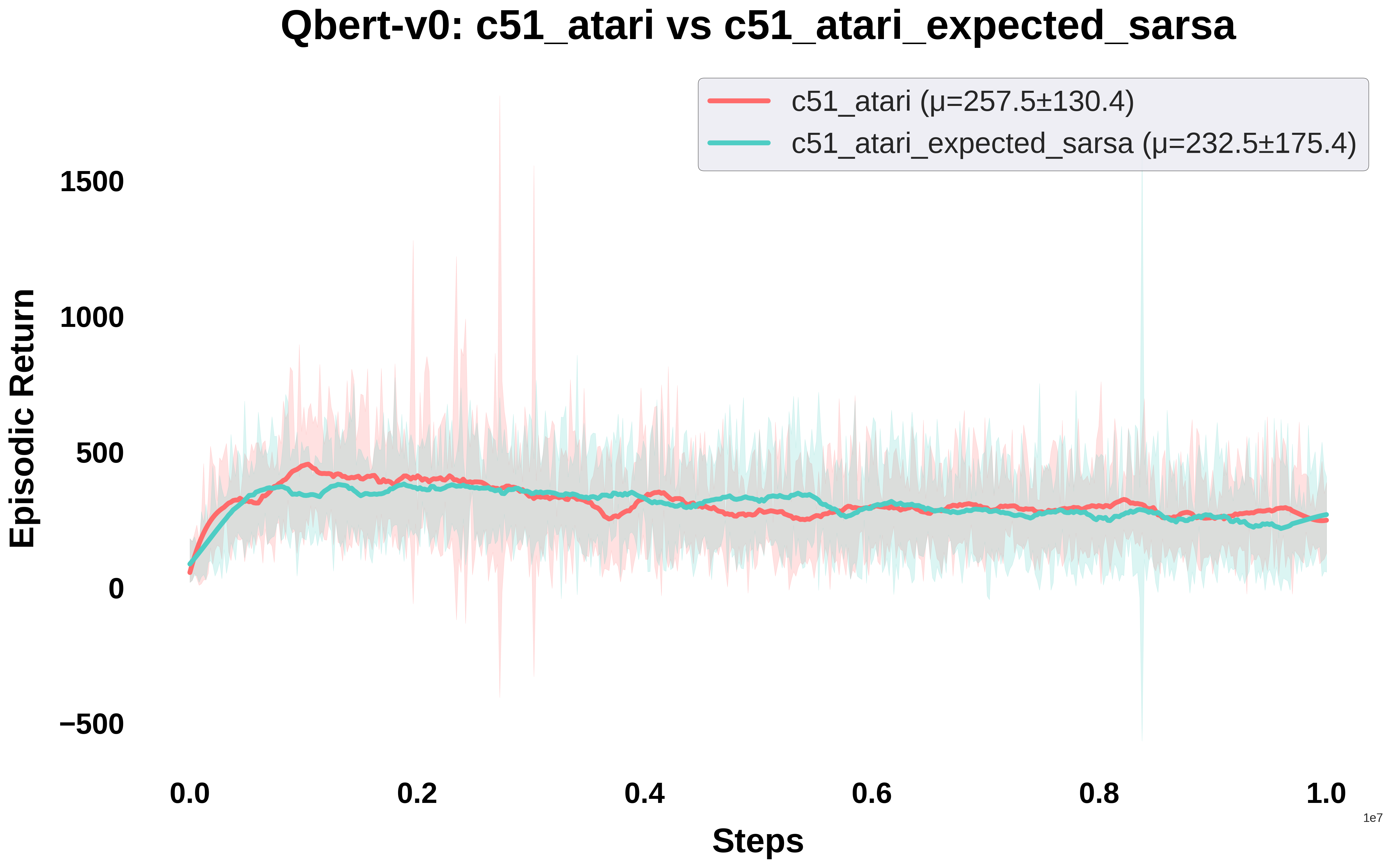}
        \caption{Qbert-v0}
        \label{fig:qbert_v0}
    \end{subfigure}
    \hfill
    \begin{subfigure}[b]{0.48\linewidth}
        \includegraphics[width=\linewidth]{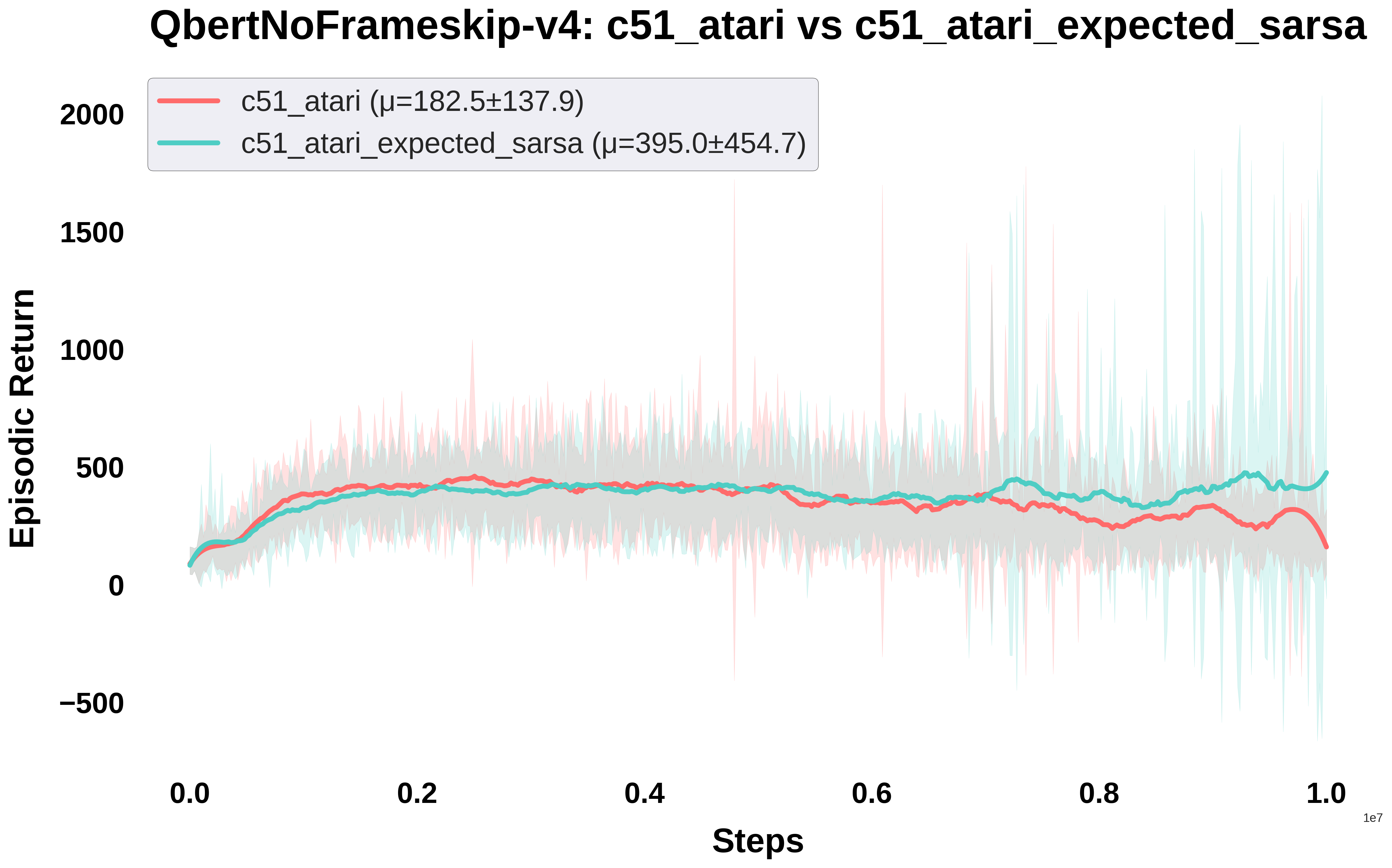}
        \caption{QbertNoFrameskip-v4}
        \label{fig:qbert_no_frameskip}
    \end{subfigure}
    \caption{Comparison results for Qbert environments.}
    \label{fig:qbert_pair}
\end{figure}

Each Atari comparison plot illustrates the performance of ES-C51 and QL-C51 on both stochastic and deterministic versions of the environments. By convention, the NoFrameskip variant represents the deterministic setting, while the v0 variant corresponds to the stochastic setting. 

Figures~\ref{fig:Amidar_pair}, \ref{fig:battlezone_pair}, \ref{fig:doubledunk_pair}, and~\ref{fig:qbert_pair} present the performance comparison of QL-C51 and ES-C51 on the \textit{Amidar}, \textit{BattleZone}, \textit{DoubleDunk}, and \textit{Qbert} environments. In all cases both algorithms exhibit suboptimal performance, In Amidar the learning curves stabilize around an average reward of approximately 20 in the stochastic version (Figure~\ref{fig:Amidar_v0}) and 5 in the deterministic version (Figure~\ref{fig:Amidar_no_frameskip}). While QL-C51 slightly outperforms ES-C51 in the stochastic variant, ES-C51 achieves marginally better results in both variants. A similar trend is observed in \textit{BattleZone}, where both algorithms plateau around an average reward of 2000 in both stochastic and deterministic versions (Figures~\ref{fig:battlezone_v0} and~\ref{fig:battlezone_no_frameskip}). For \textit{DoubleDunk}, with average rewards remaining in the negative range (around $-15$ to $-20$) across both stochastic (Figure~\ref{fig:doubledunk_v0}) and deterministic (Figure~\ref{fig:doubledunk_no_frameskip}) settings, indicating that neither approach is able to effectively adapt to the environment. In contrast, the \textit{Qbert} environment shows slightly stronger learning behavior in deterministic version (Figure~\ref{fig:qbert_no_frameskip}) where ES-C51 surpasses QL-C51 , but both have similar performance in the stochastic version (Figure~\ref{fig:qbert_v0}). Overall, across this subset of environments both algorithms perform poorly, failing to learn an optimal policy.

\begin{figure}[H]
    \centering
    \begin{subfigure}[b]{0.48\linewidth}
        \includegraphics[width=\linewidth]{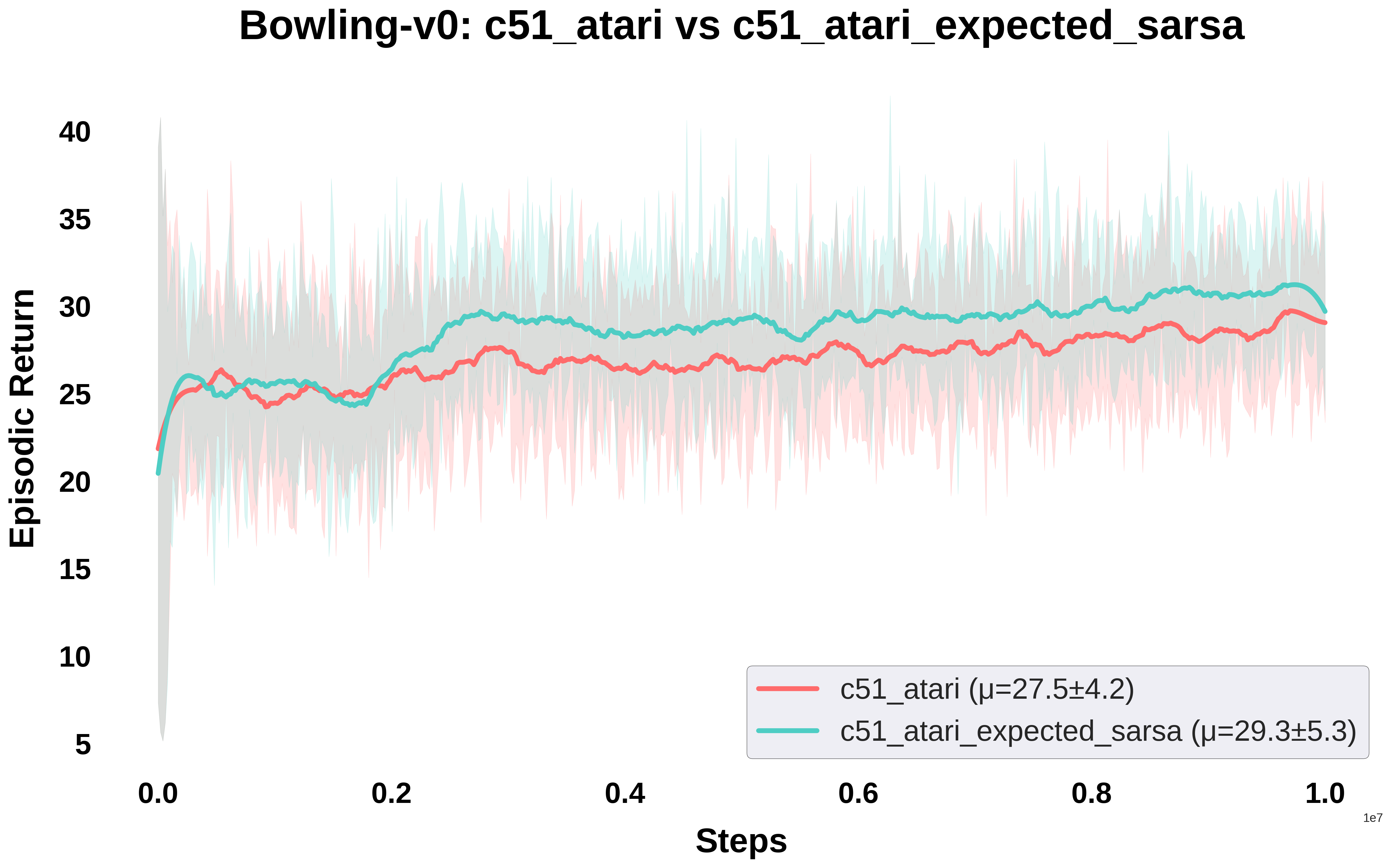}
        \caption{Bowling-v0}
        \label{fig:bowling_v0}
    \end{subfigure}
    \hfill
    \begin{subfigure}[b]{0.48\linewidth}
        \includegraphics[width=\linewidth]{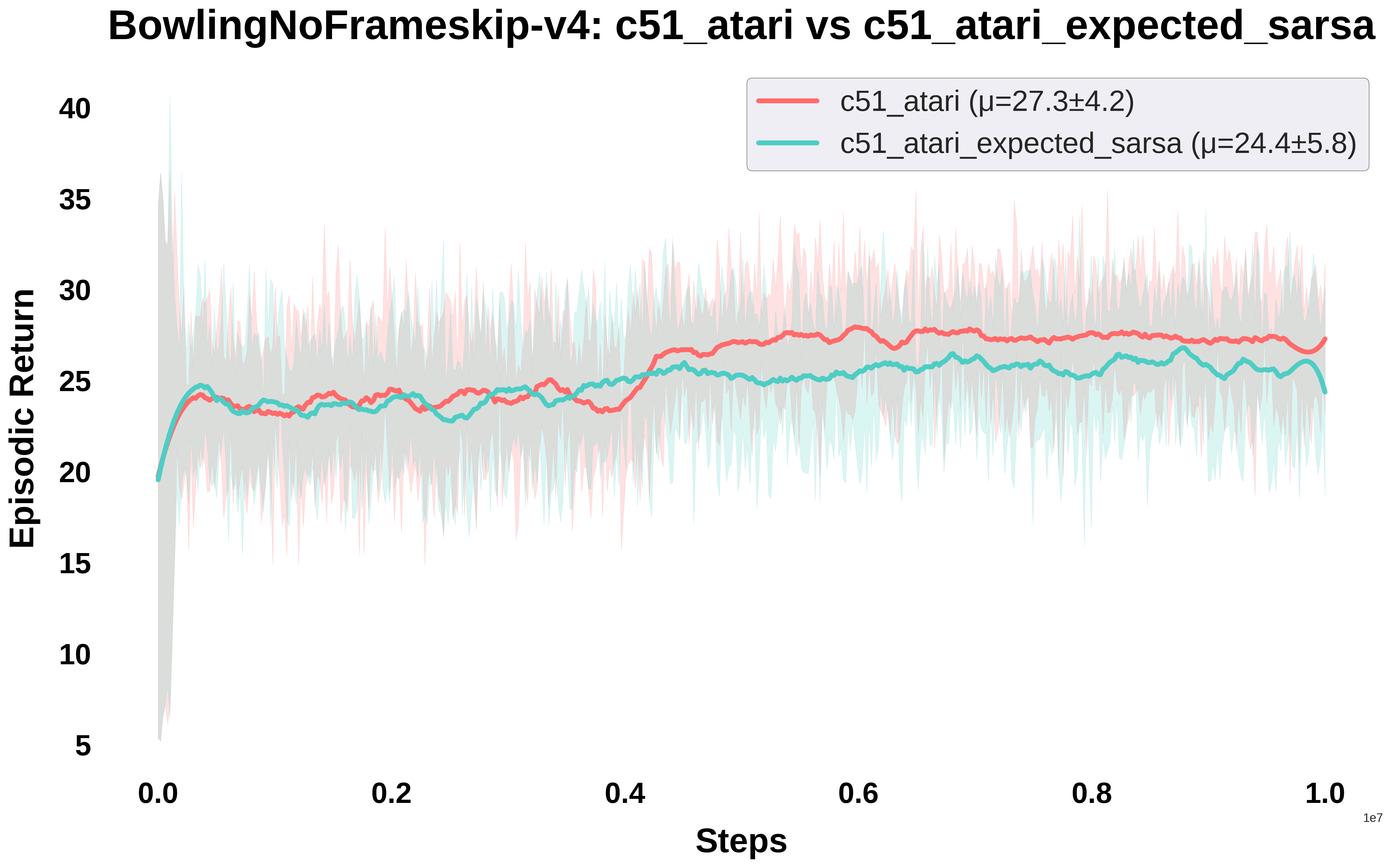}
        \caption{BowlingNoFrameskip-v4}
        \label{fig:bowling_no_frameskip}
    \end{subfigure}
    \caption{Comparison results for Bowling environments.}
    \label{fig:bowling_pair}
\end{figure}

Figure \ref{fig:bowling_pair} presents the results on the Bowling environments. In the stochastic version (\ref{fig:bowling_v0}), ES-C51 demonstrates faster convergence and achieves higher rewards, consistently surpassing QL-C51 by converging above 30, while QL-C51 struggles to reach this level. In contrast, in the deterministic version (\ref{fig:bowling_no_frameskip}), QL-C51 outperforms ES-C51, showing both higher reward convergence and more stable performance.

\begin{figure}[H]
    \centering
    \begin{subfigure}[b]{0.48\linewidth}
        \includegraphics[width=\linewidth]{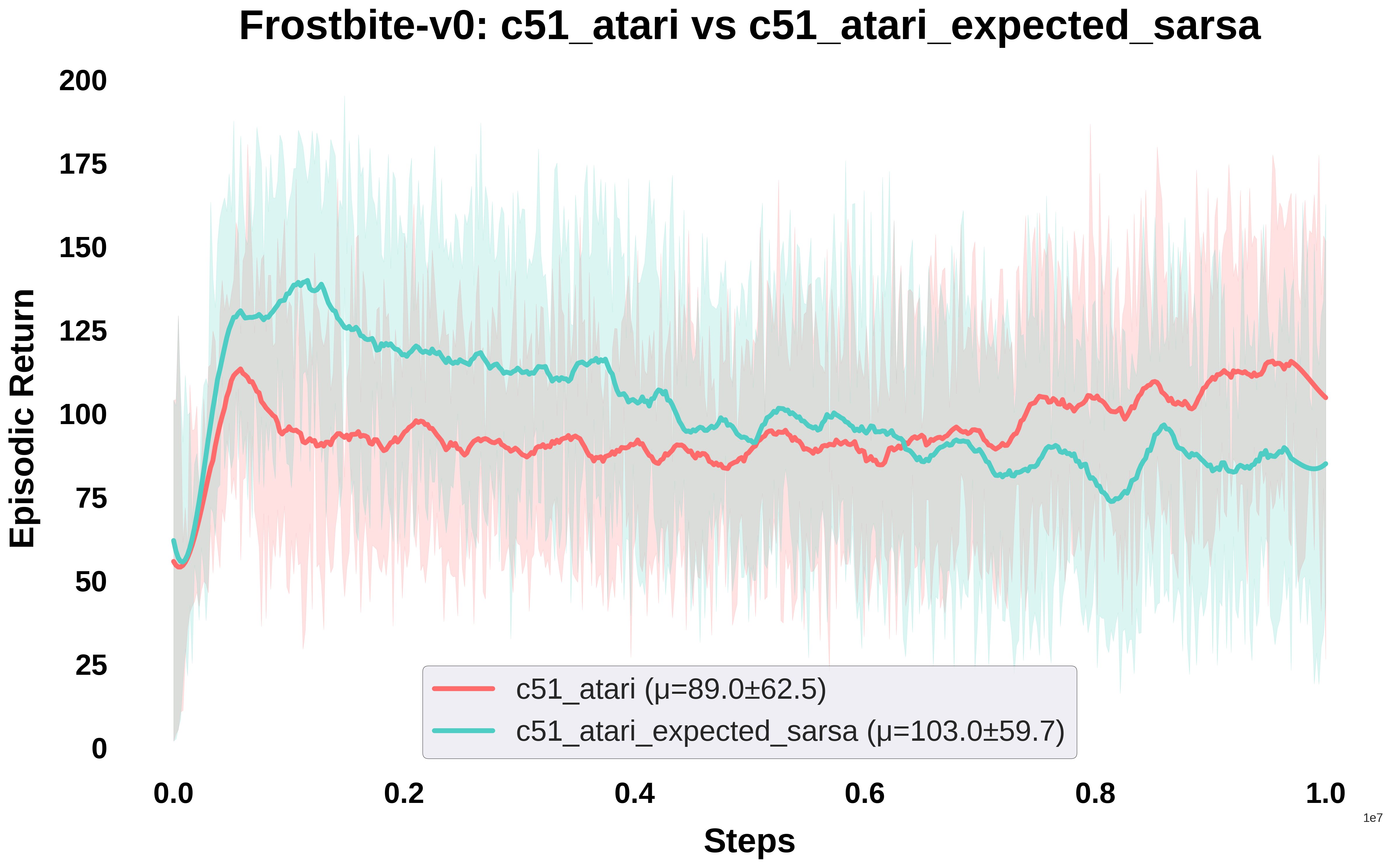}
        \caption{Frostbite-v0}
        \label{fig:frostbite_v0}
    \end{subfigure}
    \hfill
    \begin{subfigure}[b]{0.48\linewidth}
        \includegraphics[width=\linewidth]{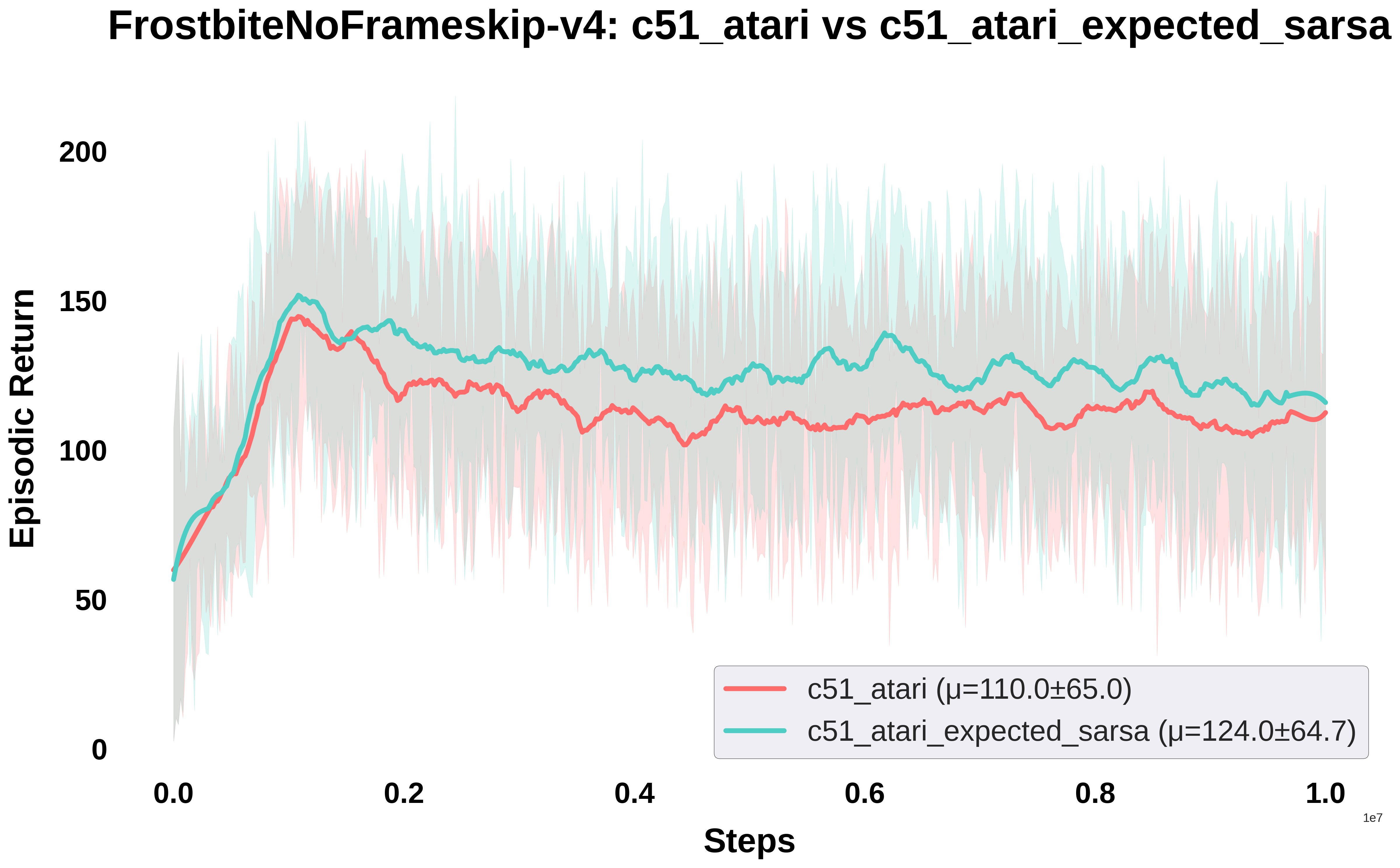}
        \caption{FrostbiteNoFrameskip-v4}
        \label{fig:frostbite_no_frameskip}
    \end{subfigure}
    \caption{Comparison results for Frostbite environments.}
    \label{fig:frostbite_pair}
\end{figure}

\begin{figure}[H]
    \centering
    \begin{subfigure}[b]{0.48\linewidth}
        \includegraphics[width=\linewidth]{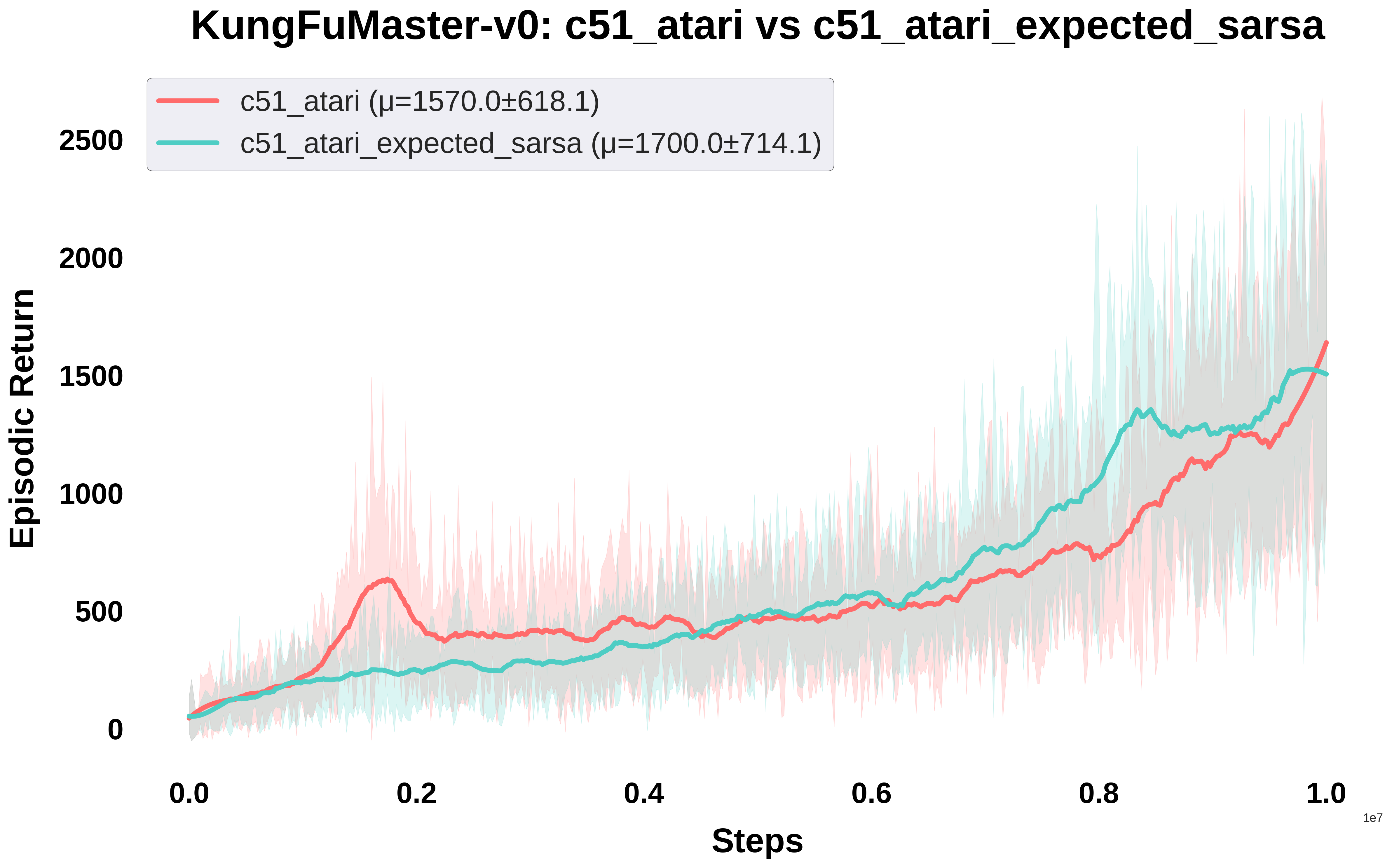}
        \caption{KungFuMaster-v0}
        \label{fig:kungfumaster_v0}
    \end{subfigure}
    \hfill
    \begin{subfigure}[b]{0.48\linewidth}
        \includegraphics[width=\linewidth]{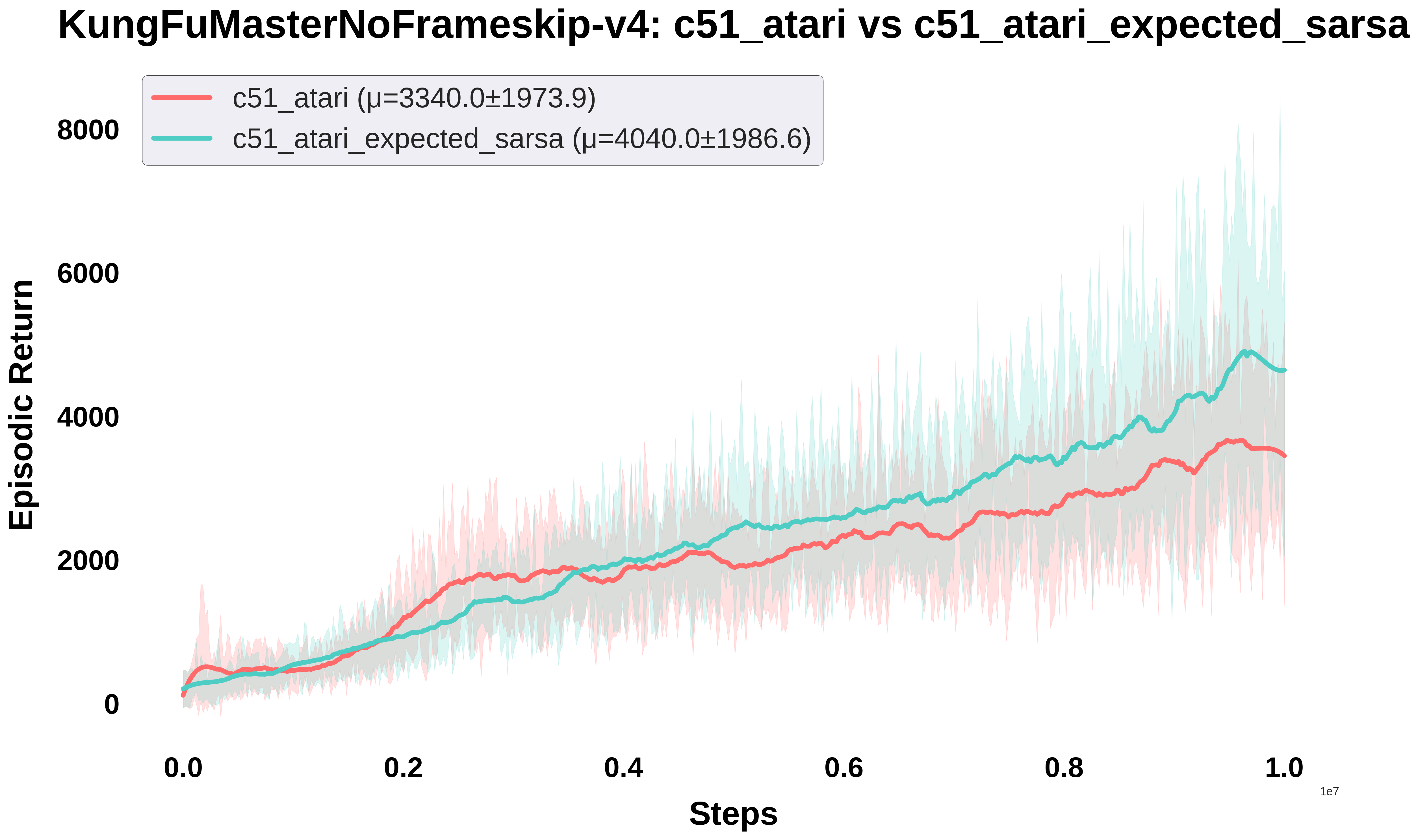}
        \caption{KungFuMasterNoFrameskip-v4}
        \label{fig:kungfumaster_no_frameskip}
    \end{subfigure}
    \caption{Comparison results for KungFuMaster environments.}
    \label{fig:kungfumaster_pair}
\end{figure}

\begin{figure}[H]
    \centering
    \begin{subfigure}[b]{0.48\linewidth}
        \includegraphics[width=\linewidth]{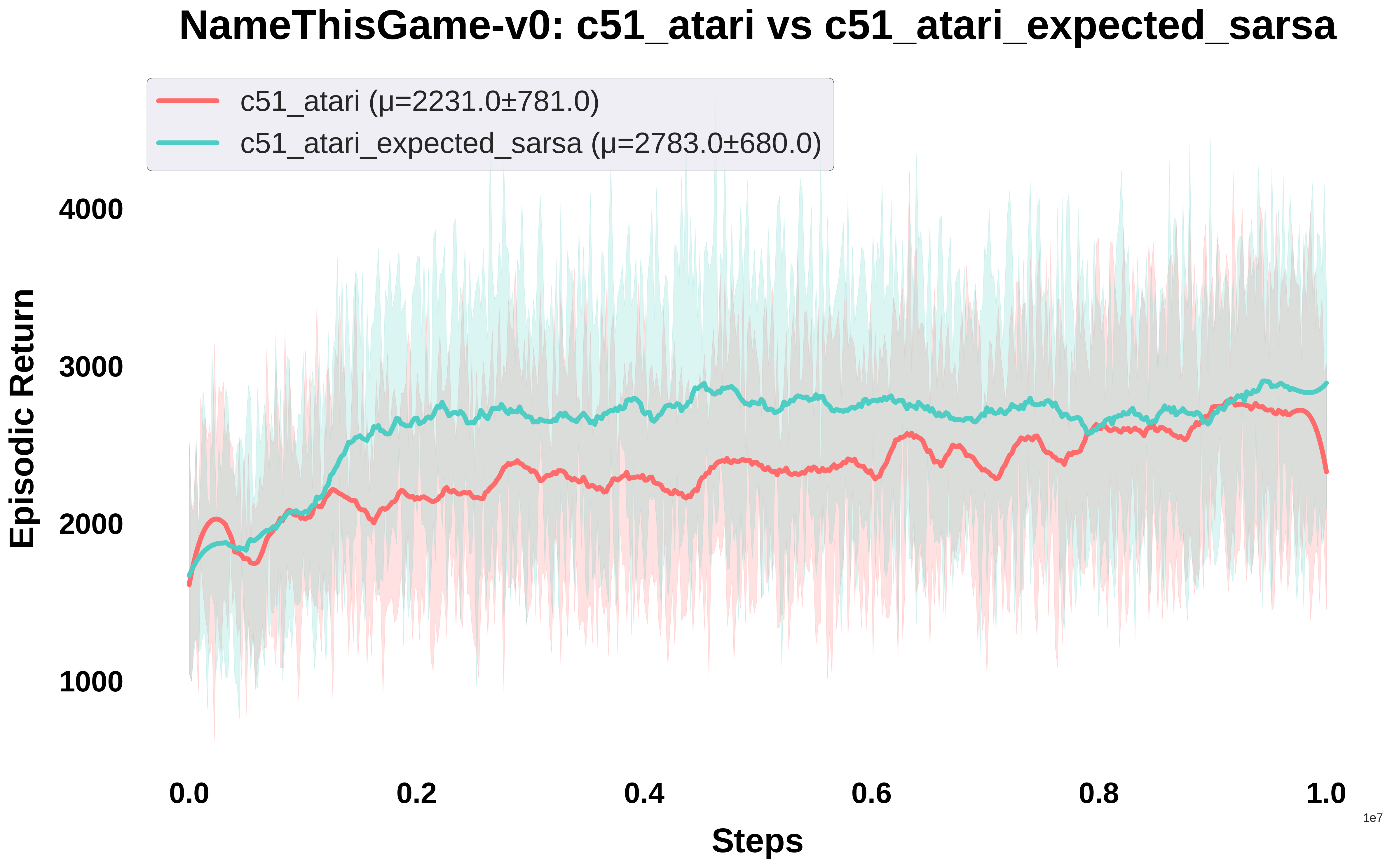}
        \caption{NameThisGame-v0}
        \label{fig:namethisgame_v0}
    \end{subfigure}
    \hfill
    \begin{subfigure}[b]{0.48\linewidth}
        \includegraphics[width=\linewidth]{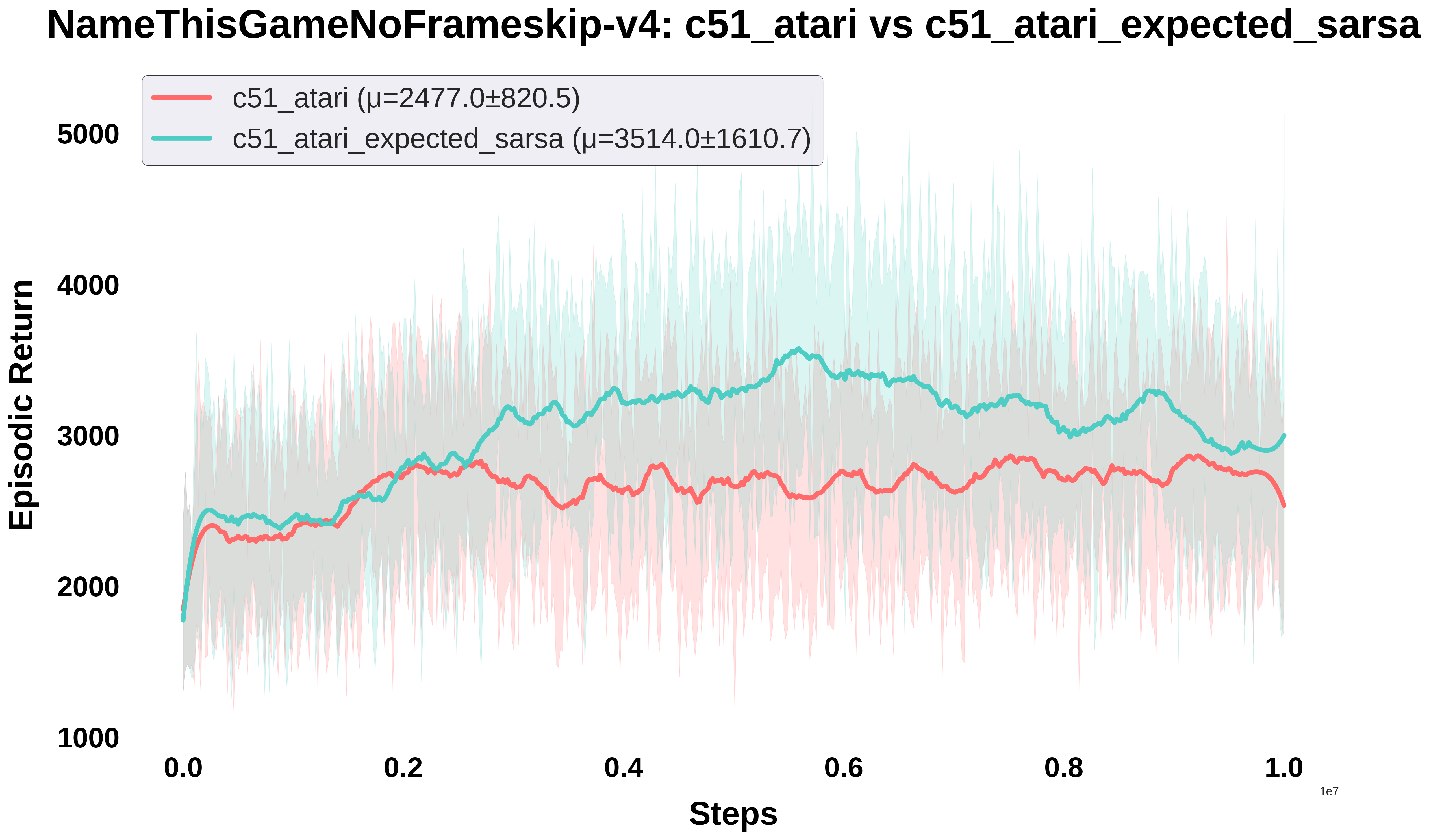}
        \caption{NameThisGameNoFrameskip-v4}
        \label{fig:namethisgame_no_frameskip}
    \end{subfigure}
    \caption{Comparison results for NameThisGame environments.}
    \label{fig:namethisgame_pair}
\end{figure}

\begin{figure}[H]
    \centering
    \begin{subfigure}[b]{0.48\linewidth}
        \includegraphics[width=\linewidth]{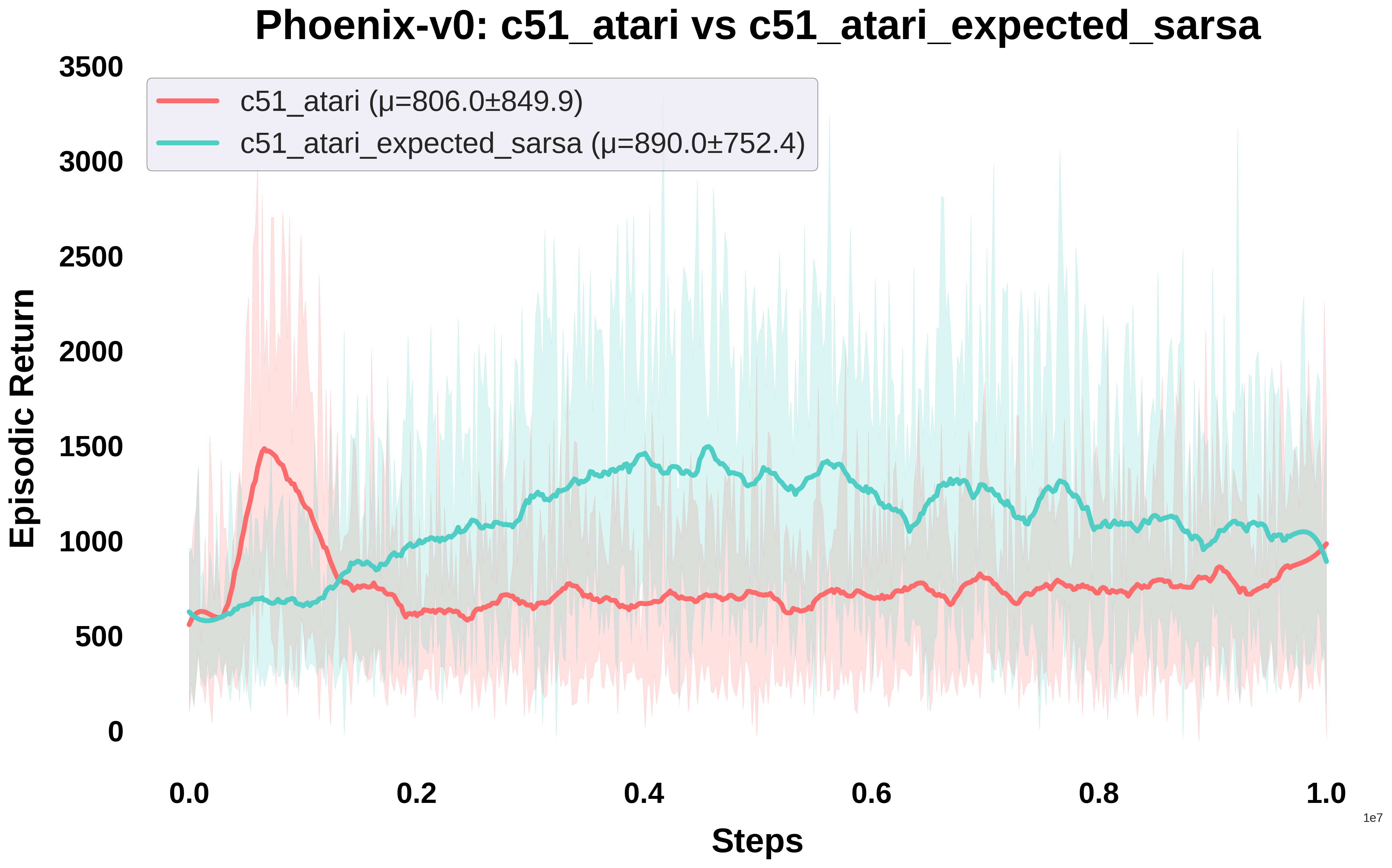}
        \caption{Phoenix-v0}
        \label{fig:phoenix_v0}
    \end{subfigure}
    \hfill
    \begin{subfigure}[b]{0.48\linewidth}
        \includegraphics[width=\linewidth]{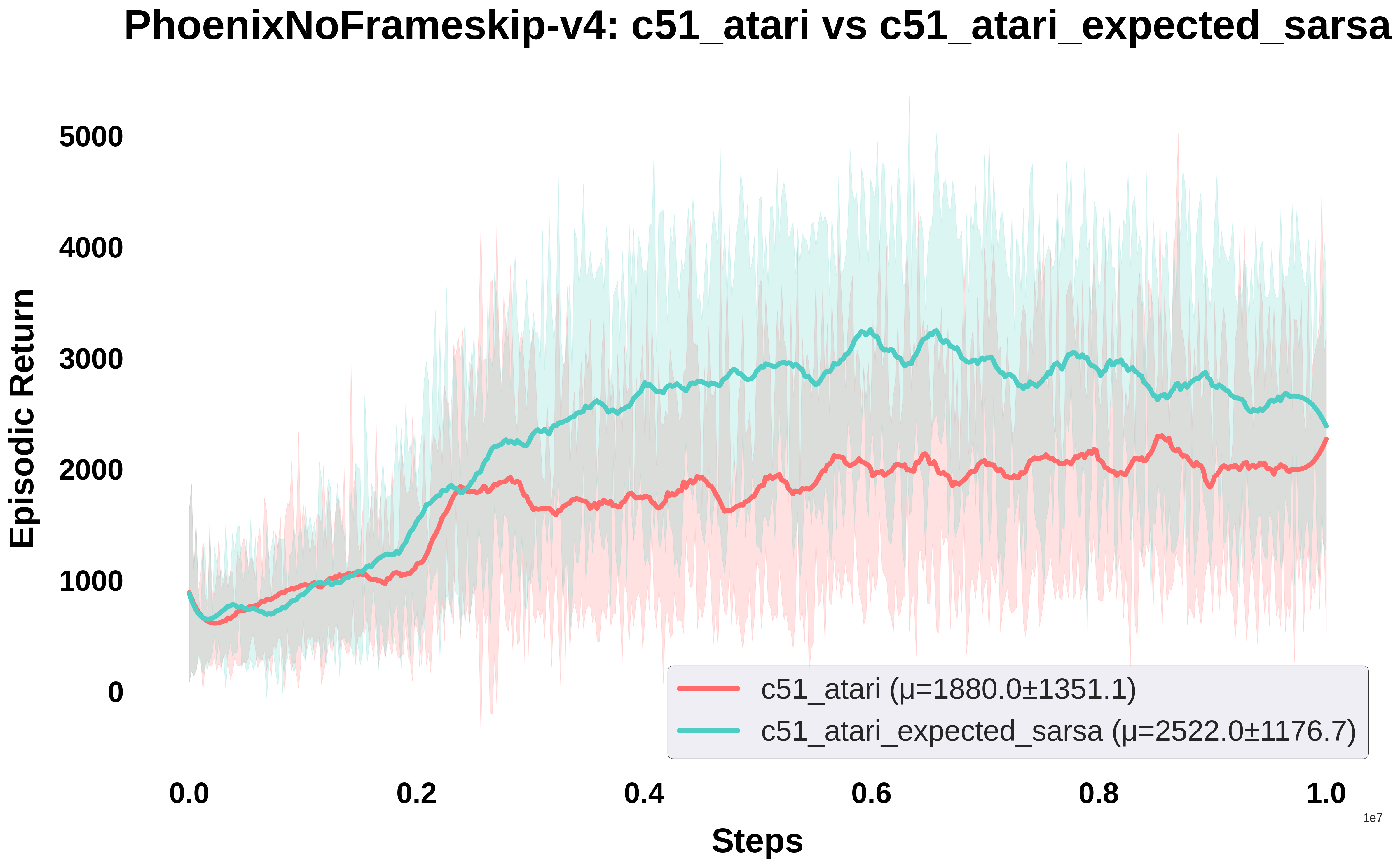}
        \caption{PhoenixNoFrameskip-v4}
        \label{fig:phoenix_no_frameskip}
    \end{subfigure}
    \caption{Comparison results for Phoenix environments.}
    \label{fig:phoenix_pair}
\end{figure}

\begin{figure}[H]
    \centering
    \begin{subfigure}[b]{0.48\linewidth}
        \includegraphics[width=\linewidth]{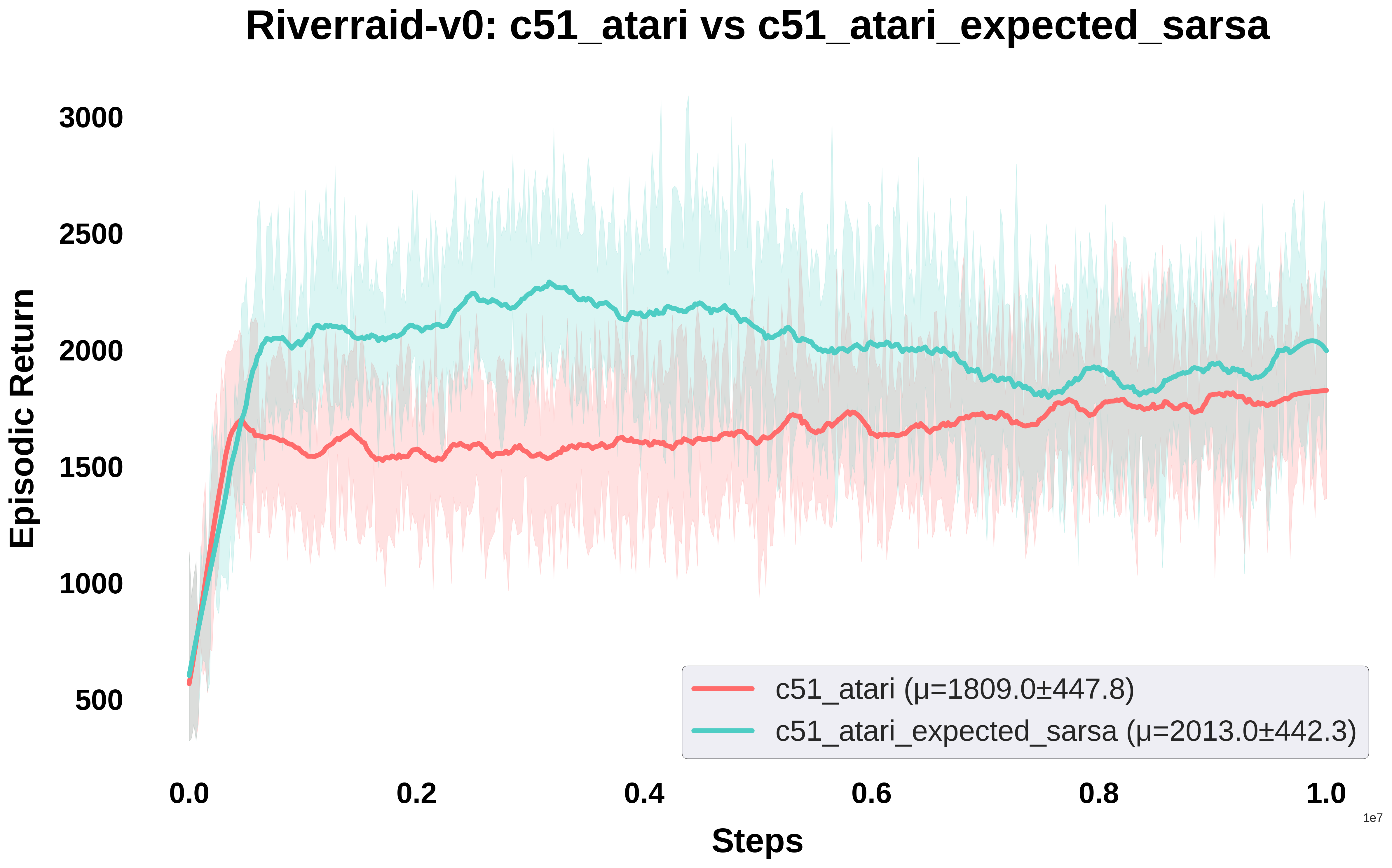}
        \caption{Riverraid-v0}
        \label{fig:riverraid_v0}
    \end{subfigure}
    \hfill
    \begin{subfigure}[b]{0.48\linewidth}
        \includegraphics[width=\linewidth]{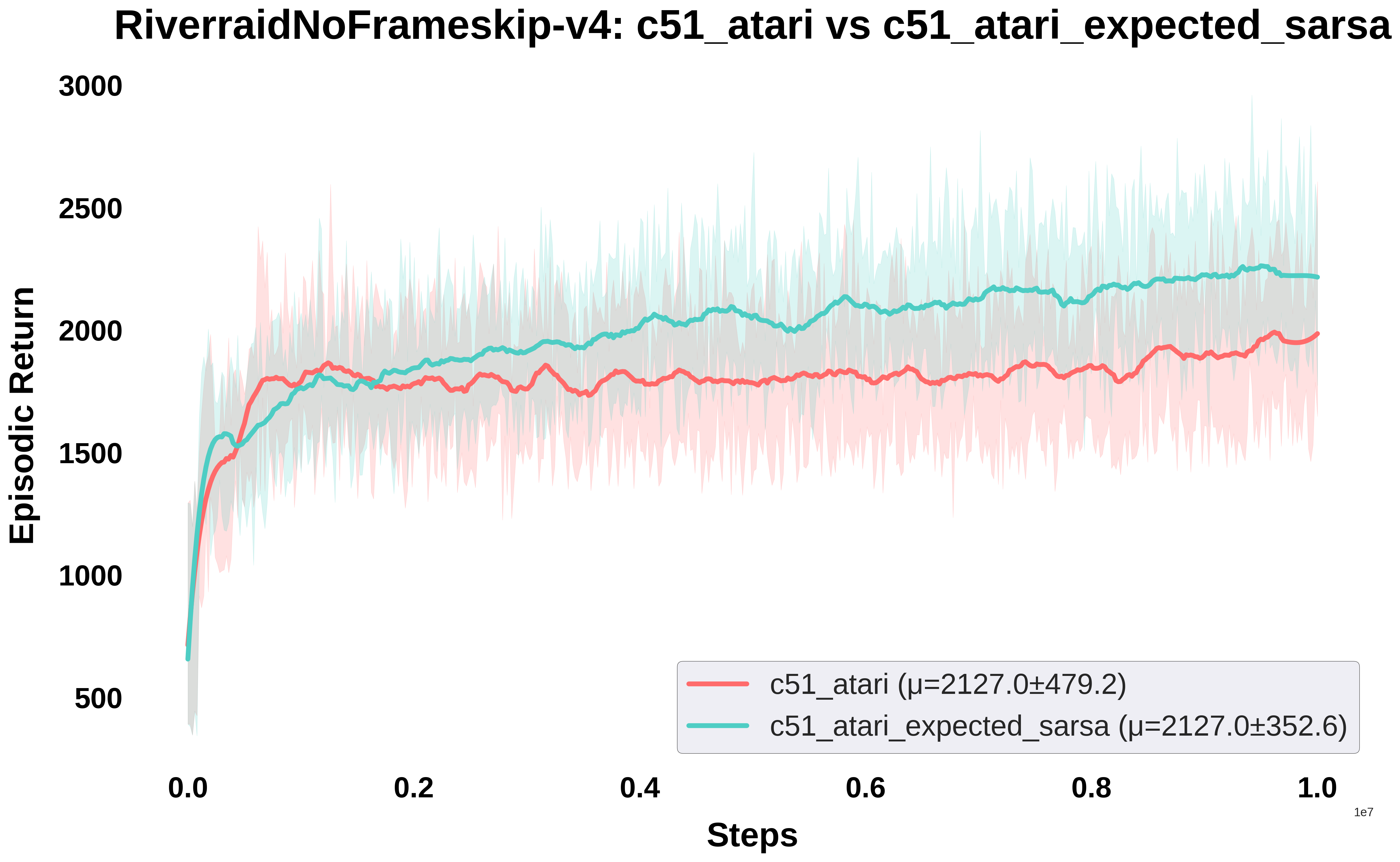}
        \caption{RiverraidNoFrameskip-v4}
        \label{fig:riverraid_no_frameskip}
    \end{subfigure}
    \caption{Comparison results for Riverraid environments.}
    \label{fig:riverraid_pair}
\end{figure}

Figures~\ref{fig:frostbite_pair}, \ref{fig:kungfumaster_pair}, \ref{fig:namethisgame_pair}, \ref{fig:phoenix_pair}, and~\ref{fig:riverraid_pair} present the comparative performance of QL-C51 and ES-C51 across the \textit{Frostbite}, \textit{KungFuMaster}, \textit{NameThisGame}, \textit{Phoenix}, and \textit{Riverraid} environments.

In the stochastic versions, a common trend is observed: ES-C51 initially achieves a sharp positive reward gain, but then drops near the end of an episode. In stochastic environments, the optimal action at a state $S_t$ is expected to change, so greedy selection should be avoided until all time steps of an episode are completed. This behavior is managed by the parameter $\tau$ in the softmax policy.

However, this early advantage gradually decreases over time. The only notable exception is \textit{KungFuMaster} \ref{fig:kungfumaster_v0} where the reward increases continuously . In general in these stochastic environments ES-C51 maintains its dominance, consistently converging to higher reward levels compared to QL-C51 (Figures~\ref{fig:kungfumaster_v0}, \ref{fig:namethisgame_v0}, \ref{fig:phoenix_v0}, \ref{fig:riverraid_v0}), With the only exception being (\ref{fig:frostbite_v0}), where ES-C51’s reward steadily declines after the initial surge, ultimately allowing QL-C51 to surpass it at around 0.7 million steps.

In the deterministic versions ES-C51 exhibits a clear and consistent advantage across all environments. Unlike in the stochastic case, its performance remains stable and continues to improve throughout training, significantly outperforming QL-C51 in \textit{Frostbite}, \textit{KungFuMaster}, \textit{NameThisGame}, \textit{Phoenix}, and \textit{Riverraid} (Figures~\ref{fig:frostbite_no_frameskip}, \ref{fig:kungfumaster_no_frameskip}, \ref{fig:namethisgame_no_frameskip}, \ref{fig:phoenix_no_frameskip}, \ref{fig:riverraid_no_frameskip}). 

We found that the performance of all algorithms is sensitive to both the value and decay of $\tau$. Different $\tau$ settings lead to noticeable variations in performance across all environments. This indicates a broad opportunity for future work to determine the most optimal initial $\tau$ value and decay rate for each algorithm–environment pair.  

Overall, ES-C51 frequently outperforms QL-C51 across both categories of environment. In the stochastic Atari environments, ES-C51 demonstrates a clear advantage, maintaining significantly higher performance up to around 0.75 million steps. Beyond this point, as $\tau$ decays to zero and the policy becomes fully greedy, ES-C51 occasionally loses its edge and experiences a performance drop in certain environments. This drawback can be alleviated by selecting the most optimal value of $tau$ for that environment. Such a modification would help retain a degree of exploration and sustain ES-C51’s advantage over longer horizons. In deterministic environments, ES-C51 delivers either superior or comparable performance relative to QL-C51. Notably, in six of the evaluated environments, ES-C51 achieves substantial improvements over QL-C51, further highlighting its robustness and effectiveness. 

\FloatBarrier

\section{Conclusion}
This study addressed the  limitation of greedy updates in the C51 distributional reinforcement learning algorithm and introduced a softmax based alternative ES-C51. Through extensive experimentation across a diverse set of Atari environments, results demonstrate that ES-C51 in many cases achieves superior performance compared to the baseline QL-C51 algorithm. This improvement highlights the effectiveness of softmax updates in mitigating the instability introduced by greedy policies, thereby enabling more reliable value distribution learning. Importantly these performance improvements are achieved at a computational cost which is comparable to the QL-C51 algorithm. Overall, this work establishes ES-C51 as a promising and scalable advancement and provides extensive experiments and explanations in support.

\begin{figure}
    \centering
    \includegraphics[width=0.8\linewidth]{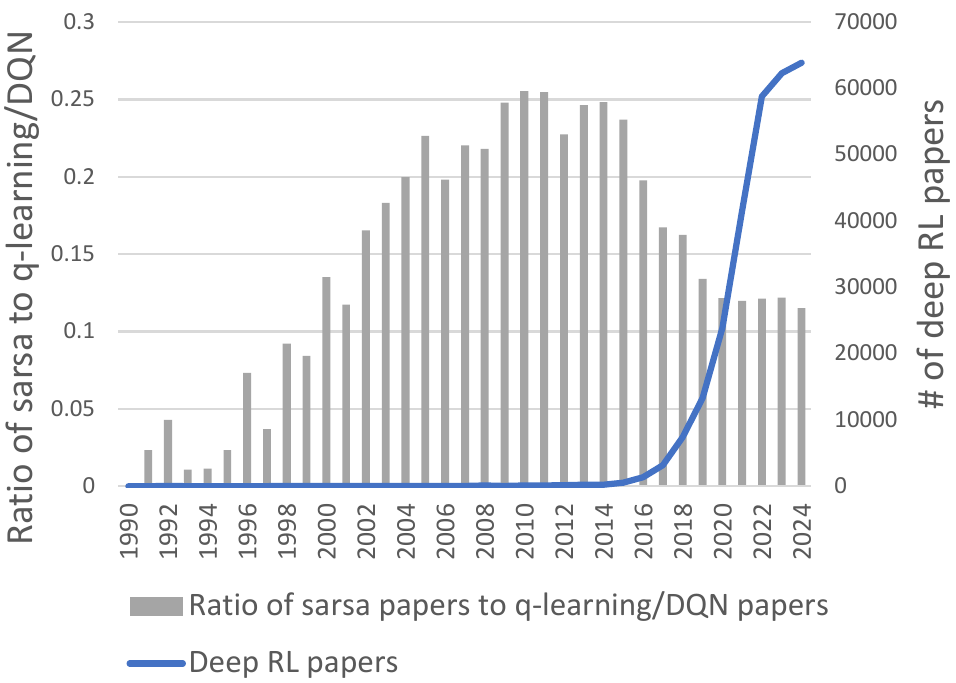}
    \caption{An analysis of publication trends in reinforcement learning, derived from Google Scholar records.}
    \label{fig:Sarsa-graph}
\end{figure}

More broadly we suggest our work offers a counterexample to a trend in reinforcement learning research over the last decade, which has seen diminished interest in on-policy methods such as Sarsa and Expected Sarsa relative to off-policy approaches like Q-learning. Figure \ref {fig:Sarsa-graph} shows an analysis of reinforcement learning publications since 1990 (statistics derived from Google Scholar). The two decades from 1990 to 2010 saw a steady increase in the use of Sarsa compared to Q-learning - while the latter was still the dominant value-based RL approach, Sarsa had grown in popularity to be mentioned in around 25\% as many papers as Q-learning. However Sarsa's usage has waned dramatically since 2015, falling by a factor of more than two relative to Q-learning. This period coincides with the boom in deep RL. Our conjecture is that the early successes of DQN encouraged the RL research community to become overly focused on deep variants of Q-learning, largely overlooking the potential for deep extensions of Sarsa and Expected Sarsa.

The results presented in this paper provide evidence that on-policy approaches such as Expected Sarsa are compatible with deep RL, and that they may provide benefits over more widely used off-policy approaches. We hope our work may spark renewed interest in on-policy value-based deep RL. 

\section*{Data availability}
The dataset of results for individual runs of each algorithm on each environment is available from \url{https://federation.figshare.com/articles/dataset/Dataset_from_ES-C51_Expected_Sarsa_Based_C51_Distributional_Reinforcement_Learning_Algorithm/30359872}, and the code implementations of ES-C51 and QL-C51 are available from
\url{https://github.com/Rijul-Tandon/cleanrl_ES-C51.git}

\section*{Acknowledgements}
The authors acknowledge the use of a large language model (LLM) to assist with paraphrasing and code generation for data visualization. The authors declare that this research did not receive any specific grant from funding agencies in the public, commercial, or not-for-profit sectors.
\FloatBarrier
\bibliographystyle{elsarticle-num}
\bibliography{ref}

\FloatBarrier

\end{document}